\pdfoutput=1
\documentclass[11pt]{article}

\usepackage{ACL2023}

\usepackage{graphicx}
\graphicspath{{./Images/}}
\usepackage{multirow}
\usepackage{comment}
\usepackage{hyperref}
\usepackage{amsmath}
\usepackage{amssymb}
\usepackage{makecell}
\usepackage{listings}

\usepackage{times}
\usepackage{latexsym}

\usepackage[T1]{fontenc}
\usepackage[utf8]{inputenc}

\usepackage{microtype}

\usepackage{inconsolata}

\title{Can Large Language Models Be an Alternative to Human Evaluation?}

\author{Cheng-Han Chiang \\
  National Taiwan University,\\ Taiwan\\
  \texttt{dcml0714@gmail.com} \\\And
   Hung-yi Lee \\
  National Taiwan University,\\ Taiwan \\
  \texttt{hungyilee@ntu.edu.tw} \\}

\begin{document}
\maketitle
\begin{abstract}
Human evaluation is indispensable and inevitable for assessing the quality of texts generated by machine learning models or written by humans. 
However, human evaluation is very difficult to reproduce and its quality is notoriously unstable, hindering fair comparisons among different natural language processing (NLP) models and algorithms.
Recently, large language models (LLMs) have demonstrated exceptional performance on unseen tasks when only the task instructions are provided.
In this paper, we explore if such an ability of the LLMs can be used as an alternative to human evaluation.
We present the LLMs with the exact same instructions, samples to be evaluated, and questions used to conduct human evaluation, and then ask the LLMs to generate responses to those questions; we dub this \textit{LLM evaluation}.
We use human evaluation and LLM evaluation to evaluate the texts in two NLP tasks: open-ended story generation and adversarial attacks.
We show that the result of LLM evaluation is consistent with the results obtained by expert human evaluation: the texts rated higher by human experts are also rated higher by the LLMs.
We also find that the results of LLM evaluation are stable over different formatting of the task instructions and the sampling algorithm used to generate the answer.
We are the first to show the potential of using LLMs to assess the quality of texts and discuss the limitations and ethical considerations of LLM evaluation.

\end{abstract}

\section{Introduction}
Human evaluation is an important method to understand the performance of an NLP model or algorithm~\citep{guzman-etal-2015-humans, gillick-liu-2010-non}.
We rely on human evaluation because there are certain aspects of texts that are hard to evaluate using automatic evaluation metrics; thus, researchers resort to humans to rate the quality of the output of NLP models.
While human evaluation is prevalent and indispensable in NLP, it is notoriously unstable~\citep{gillick-liu-2010-non, clark-etal-2021-thats}. 
\citet{karpinska-etal-2021-perils} has shown that low-quality workforces in human evaluation can have a detrimental effect on the evaluation result, making it impossible to compare the performance among different systems. 
Reproducibility is another issue in human evaluation since it is hard to recruit the same human evaluators and rerun the same evaluation. 
Even if the same workers are recruited, the workers that have seen the task before are likely to produce a different evaluation result the next time because they have already done the task.
While human evaluation is used to better assess NLP systems and has some advantages over automatic evaluation metrics, the drawbacks of human evaluation somewhat make it difficult to reliably evaluate NLP systems.

\begin{figure*}[t]
\centering
\includegraphics[clip, trim = 0px 0px 0px 00px,width=0.95\linewidth]{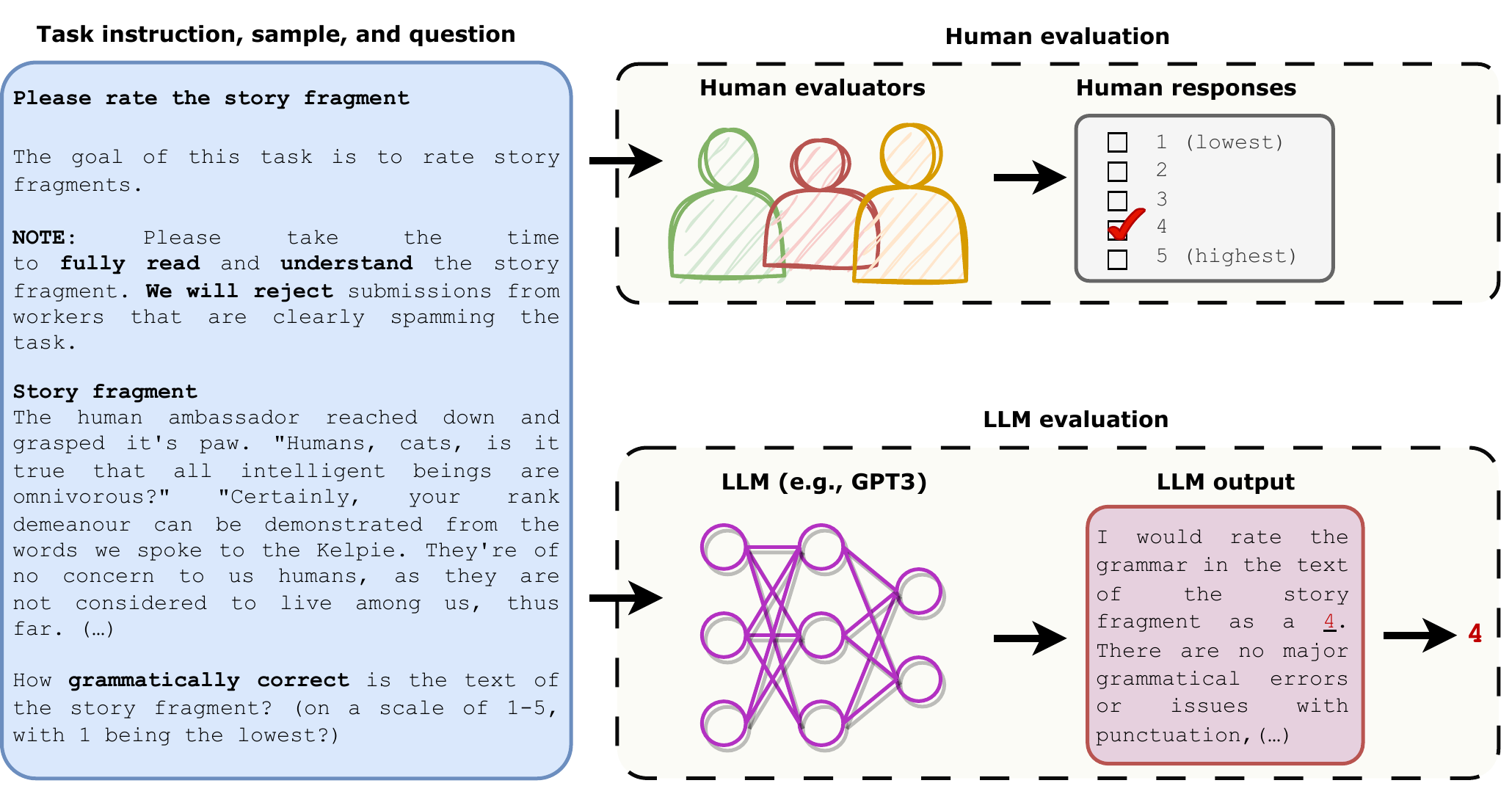}
\caption{Illustration of the core idea of the paper using open-ended story generation as the example task.
The left part shows the instruction, story fragments, and questions used in human evaluation.
The human experts are asked to rate the quality of the story fragments using a 5-point Likert scale, shown on the upper right.
The lower right part shows the process of \textit{LLM evaluation}, where we feed the LLMs the same instruction, story fragments, and questions and parse the LLM-generated output to get the rating.
}
\label{fig:Illustration.pdf}
\end{figure*}

To resolve some of the drawbacks, we take advantage of large language models (LLMs). 
LLMs are large models that are trained to model human languages using self-supervised learning~\citep{brown2020language} and further using special training procedures to improve the performance on unseen tasks and better follow natural language instructions~\citep{sanh2022multitask,wei2022finetuned}.
The ability to perform a task just given the task instructions motivates us to ask if these LLMs can perform what humans do in human evaluation.
To answer this question, we feed in the LLM with the same instruction, sample, and question used in human evaluation, and take the sequences generated by the LLM as the LLM's answer to the question.
This process is shown in Figure~\ref{fig:Illustration.pdf}, and we call this process \textbf{LLM evaluation}.

To test if LLM evaluation yields meaningful results, we conduct LLM evaluation on two different NLP tasks: evaluating the quality of stories in open-ended story generation and the quality of sentences generated by adversarial attacks.
We summarize our findings and contribution as follows:
\begin{itemize}
    \item We show that LLM evaluation produces results similar to expert human evaluation, verifying the effectiveness of LLM evaluation
    (\S\ref{subsection: task 1 Experiment Results} and \S\ref{subsection: task 2 result}).
    This paper is \textbf{the first} to propose using LLMs as an alternative to human evaluation and show their effectiveness.
    \item We show that LLM evaluation results only slightly vary due to different task instructions and the hyperparameters of the sampling algorithm  used to generate the answer. (\S\ref{seubsubsection: Variance due to Different Instructions} and \S\ref{subsubsection: Variance due to Different Sampling Parameters})
    \item We carefully discuss the pros and cons of using LLM evaluation and discuss the ethical considerations of LLM evaluation. (\S\ref{section: discussions}) 
\end{itemize}

\section{LLM Evaluation}
\label{section: LLM Evaluation}
\subsection{Large Language Models (LLMs)}
\label{subsection: Large Language Models (LLMs)}
Large language models are language models having bulk parameter sizes, typically on the scale of a few billion, and pre-trained on enormous amounts of natural language corpora, including GPT3~\citep{brown2020language}, T5~\citep{raffel2020exploring}, and BLOOM~\citep{scao2022bloom}.
These LLMs show exceptional performance on unseen tasks when only the task instructions are given; this kind of ability is called \textbf{zero-shot in-context learning}.

To further improve the zero-shot in-context learning performance, special training techniques have been applied to those LLMs after pre-training.
For example, T0~\citep{sanh2022multitask} and FLAN~\citep{wei2022finetuned} are fine-tuned on a mixture of tasks and can thus achieve better zero-shot performance compared to GPT-3.
InstructGPT~\citep{ouyang2022training} is fine-tuned from GPT-3 using reinforcement learning from human feedback (RLHF), and it is shown to better follow the instructions.
ChatGPT~\citep{ChatGPT} is fine-tuned from InstructGPT with a conversation dataset using RLHF, so ChatGPT can interact with users in a conversational way.
ChatGPT is able to answer questions asked by the user and provide comprehensive explanations about its answer.
%ChatGPT has been claimed to pass the Turing test~\citep{TuringChatGPT} and its phenomenal success has created a boom among NLP researchers and the general public.
%Given the remarkable ability to follow task instructions and provide feedback, we ask if such an ability can be harnessed as an alternative to human evaluation and better help NLP researchers evaluate NLP systems.
Given the LLMs' ability to follow task instructions and provide feedback, we ask whether LLMs can be used as an alternative to human evaluation and aid NLP researchers in evaluating the quality of texts.

\subsection{LLM Evaluation}
\label{subsection: LLM Evaluation}
To evaluate the quality of texts generated by NLP systems or written by humans using LLM, we present the LLMs with the task instructions, the sample to be evaluated, and a question. 
The question asks the LLM to rate the sample's quality using a 5-point Likert scale.
Given the inputs, the LLM will answer the question by generating some output sentences.
We parse the output sentences to get the score rated by the LLM.
We call this process \textit{LLM evaluation}, and this procedure is shown in the lower part of Figure~\ref{fig:Illustration.pdf}.
Different tasks use different sets of task instructions, and each task uses different questions to evaluate the quality of the samples.
The instructions and questions used in LLM evaluation in our paper are not tailored for the LLMs; we follow those instructions used to conduct human evaluation in prior works.

To compare the result of LLM evaluation and show its effectiveness, we compare the result of LLM evaluation with human evaluation conducted by English teachers.
To make a fair and meaningful comparison, the instructions, samples, and questions  in human evaluation are formatted similarly to those in LLM evaluation.
The main difference between LLM evaluation and human evaluation is that in human evaluation, the human evaluators answer the question by choosing the answer from a pre-defined set of options (the 1-5 Likert scale scores), as shown in the upper right in Figure~\ref{fig:Illustration.pdf}.
In LLM evaluation, we instead let the LLM freely generate sentences and extract the score from the generated sentences using some simple rules, detailed in Appendix~\ref{app:Parsing the LLM outputs}.

\section{Example Task 1: Open-Ended Story Generation}
\label{section: Example Task 1: Open-Ended Story Generation}
We first use open-ended story generation to demonstrate the usefulness of LLM evaluation.

\subsection{Task Introduction}
\label{subsection: task 1 introduction}
Open-ended story generation is a task to generate a short story based on a given prompt. 
We use the \texttt{WritingPrompts} dataset~\citep{fan-etal-2018-hierarchical}, which is composed of pairs of short prompts and human-written stories collected from the subreddit \href{https://www.reddit.com/r/WritingPrompts/}{\texttt{WritingPrompts}}.
In the \texttt{WritingPrompts}, the users are given a short prompt, and they need to write a story based on the short prompt.\footnote{The \texttt{WritingPrompts} subreddit explicitly forbids the users to use AI for generating stories, so we consider the stories in the dataset to be human-written.}

In this experiment, we use LLM evaluation and human evaluation to rate the stories generated by humans and the stories generated by a story generation model.
We select open-ended story generation as an example because~\citet{karpinska-etal-2021-perils} show that workers from \href{https://www.mturk.com/}{Amazon Mechanical Turk} (AMT) cannot distinguish GPT-2~\citep{radford2019language} generated and human-written stories, while English teachers show a clear preference for human-written stories over GPT-2-generated stories.
We want to see if LLM can rate human-written stories higher than GPT-2-generated ones.

Following prior works~\citep{mao-etal-2019-improving,guan2020knowledge,karpinska-etal-2021-perils}, the story generation model is GPT-2 medium model fine-tuned on the \texttt{WritingPrompts} training dataset.
After the model is trained, we randomly select 200 prompts from the testing set of \texttt{WritingPrompts} and make the fine-tuned GPT-2 generate stories based on those prompts using nucleus sampling~\citep{Holtzman2020The} with $p=0.9$.
For the human-written stories to be compared, we use the 200 stories written based on the same 200 prompts.
We post-process the human-written and GPT-2-generated stories and then use them for LLM evaluation and human evaluation.
Please find the details on fine-tuning and data processing in Appendix~\ref{section: Experiment Details for Open-Ended Story Generation}.
%The training set of \texttt{WritingPrompts} consists of 303K pairs of prompt-story pairs.

\begin{table*}[ht]
    \centering
    \begin{tabular}{ccccccccc}
       \hline
       \multirow{2}{*}{\textbf{Evaluator}} & \multicolumn{2}{c}{\underline{\textbf{Grammaticality}}}  & \multicolumn{2}{c}{\underline{\textbf{Cohesiveness}}} & \multicolumn{2}{c}{\underline{\textbf{Likability}}} & \multicolumn{2}{c}{\underline{\textbf{Relevance}}}\\
       & Mean\textsubscript{STD} & IAA\textsubscript{\%} & Mean\textsubscript{STD} & IAA\textsubscript{\%} & Mean\textsubscript{STD} & IAA\textsubscript{\%} & Mean\textsubscript{STD} & IAA\textsubscript{\%} \\
       \hline
       \hline
       \multicolumn{9}{c}{\textit{Human-written stories}} \\
       Human & $3.76_{0.95}$ & $0.33_{20.5}$ & $4.29_{0.82}$ & $0.32_{27}$ & $3.78_{1.10}$ & $0.08_{9.5}$ & $3.35_{1.48}$ & $0.05_{8}$ \\
       %\hline
       T0 & $2.55_{1.47}$& $0.16_{10}$ & $2.98_{1.45}$ & $0.11_{4}$ & $3.18_{1.53}$ & $0.12_{7}$ & $2.93_{1.64}$ & $0.02_{6}$ \\
       \texttt{curie} & $3.19_{0.47}$ & $0.07_{46.5}$ & $2.82_{0.46}$ & $0.01_{47.5}$ & $2.85_{0.37}$ & $0.11_{0.65}$ & $3.06_{0.40}$ & $0.11_{0.64}$ \\
       \texttt{davinci} & $4.22_{0.38}$ & $0.26_{35}$ & $4.54_{0.47}$ & $0.37_{39.5}$ & $3.99_{0.38}$ & $0.49_{68.5}$ & $4.40_{0.79}$ & $0.71_{48.5}$  \\
       ChatGPT& $3.83_{0.60}$& & $3.55_{0.88}$ & & $2.44_{0.89}$ & & $3.29_{1.50}$ & \\
       \hline 
       \hline
       \multicolumn{9}{c}{\textit{GPT-2-generated stories}} \\
       Human & $3.56_{0.91}$ & $0.10_{19.5}$ & $3.19_{1.07}$ & $0.14_{17}$ & $2.59_{1.29}$ & $-0.21_{3.5}$ & $2.38_{1.40}$ & $-0.03_{8.5}$ \\
       %\hline
       T0 & $2.44_{1.49}$ & $0.05_{9}$ & $3.02_{1.51}$ & $0.07_{6}$ & $3.00_{1.59}$ & $0.16_{6}$ & $2.82_{1.61}$ & $0.04_{6}$ \\
       \texttt{curie} & $3.23_{0.51}$ & $0.01_{38}$ & $2.82_{0.45}$ & $0.02_{50}$ & $2.86_{0.37}$ & $0.09_{65.5}$ & $3.01_{0.43}$ & $0.11_{61}$ \\
       \texttt{davinci} & $4.07_{0.35}$ & $0.35_{45.5}$ & $4.26_{0.45}$ & $0.42_{42}$ & $3.84_{0.42}$ & $0.52_{62}$ & $4.02_{0.74}$ & $0.69_{42.5}$ \\
       ChatGPT & $2.98_{0.76}$ & & $2.48_{0.71}$ & & $1.59_{0.67}$ & & $2.02_{1.21}$ &\\
       \hline
    \end{tabular}
    \caption{LLM evaluation and human evaluation results of human-written stories and GPT-2-generated stories.
    For each evaluated attribute, we report its mean Likert scale and the standard deviation.
    We also report the inter-annotator agreement (IAA) among three annotators using Krippendorff’s $\alpha$.
    The subscript in the IAA column (\%) is used to denote the percentage of the stories where all three annotators exactly agree on a rating.
    }
    \label{tab: task 1 main result}
\end{table*}

\subsection{LLM Evaluation and Human Evaluation}
\label{subsection: task 1 LLM Evaluation and Human Evaluation}
We present the LLMs and the human evaluators with a short description, and the story to be evaluated, formatted as shown in Figure~\ref{fig:Illustration.pdf}.
Following~\citet{karpinska-etal-2021-perils}, we evaluate the stories on four different attributes.
The four attributes and their corresponding questions are as follows:
\begin{enumerate}
    \item \textit{Grammaticality}: How \textbf{grammatically correct} is the text of the story fragment?
    \item \textit{Cohesiveness}: How well do \textbf{the sentences} in the story fragment \textbf{fit together}?
    \item \textit{Likability}: How \textbf{enjoyable} do you find the story fragment?
    \item \textit{Relevance}: Now read the PROMPT based on which the story fragment was written. \\
    \textbf{Prompt}: \texttt{[PROMPT]}.\\
    How \textbf{relevant} is the \textbf{story fragment} to the \textbf{prompt?}
\end{enumerate}
Where the \texttt{[PROMPT]} will be filled in with the prompt which the story is based on.
Each attribute is evaluated using a 5-point Likert scale; the following description is appended at the end of each question: "\textit{(on a scale of 1-5, with 1 being the lowest)}".
We show the interface used in human evaluation and the input format for the LLM evaluation in Appendix~\ref{subsection: task 1 Human Evaluation Interfaces} and~\ref{app: task 1 instructions}.

The LLMs used for LLM evaluation include T0, \texttt{text-curie-001}, \texttt{text-davinci-003}, and ChatGPT.
\texttt{text-curie-001} and \texttt{text-davinci-003} are two InstructGPT models, and the latter is the stronger model; we will use InstructGPT to refer to these two models.
We query the InstructGPT using the official API provided by OpenAI.
We use nucleus sampling with $p=0.9$ to generate the answer from T0 and InstructGPTs.
We \textbf{sample three answers} from LLMs to stimulate the result of asking the model to rate the same story three times.
We query ChatGPT using the user interface recently released by OpenAI.
Unlike InstructGPT, we cannot control the parameters used for generating the response from ChatGPT.
Because ChatGPT limits the maximum number of queries per user, we only sample one response for each question.

For human evaluation, we do not use the commonly used AMT for human evaluation because~\citet{karpinska-etal-2021-perils} has already shown that the results obtained using AMT are highly questionable.
Following the recommendation of the prior works, we hire \textbf{three certified English teachers} using an online freelancer platform, \href{https://www.upwork.com/}{UpWork}.
Teachers are familiar with evaluating the essays of students, making them the expert evaluators in our task.
%Each English teacher rates 200 human-written stories and 200 GPT-2-generated stories.
The details about recruiting human evaluators are in Appendix~\ref{subsubsection: Hiring English Teachers and Payment}.
Each LLM and each English teacher rates the 200 human-written stories and 200 GPT-2-generated stories.

\subsection{Experiment Results}
\label{subsection: task 1 Experiment Results}
The LLM evaluation and human evaluation results of open-ended story generation are presented in Table~\ref{tab: task 1 main result}.
We report the mean and standard deviation of the Likert scores obtained from LLM evaluation and human evaluation and show the inter-annotator agreement (IAA) using two different metrics: (1) the Krippendorff’s $\alpha$, and (2) the percentage of the stories where three evaluators give the exact same rating.\footnote{The three evaluators in human evaluation are the three English teachers. In LLM evaluation, we sample the answer generated by LLM three times as an analogy to three different evaluators.}
The main observations from Table~\ref{tab: task 1 main result} are discussed as follows.

\textbf{Expert human evaluators prefer human-written stories:}
Human evaluation result serves as some kind of \textit{ground truth} of the LLM evaluation. 
For all four attributes, teachers rate the human-written stories higher than GPT-2-generated stories. 
This indicates that experts are able to distinguish the quality difference between model-generated stories and human-written stories. 
Based on the IAA, we also find that the agreements among experts are lower on GPT-2-generated texts and on the \textit{likability}.
This shows that experts tend to have less agreement on model-generated texts and on a subjective attribute (\textit{likability}), agreeing with the results in~\citet{karpinska-etal-2021-perils}.

%Next, we turn our attention to the results of LLM evaluation.

\textbf{T0 and \texttt{text-curie-001} do not show clear preference toward human-written stories}:
For T0, we can see that T0 rates human-written stories higher than GPT-2-generated stories on grammatically, likability, and relevance. 
However, the rating differences between the human-written and model-generated stories do not achieve statistical significance for \textit{grammaticality} and \textit{relevance}; the \textit{p}-value obtained by Welch's \textit{t}-test is much larger than $0.05$.
The result of \texttt{text-curie-001} is similar to T0: \texttt{text-curie-001} do not rate human-written stories higher than model-generated stories.
It can also be observed that for T0, the IAA in terms of the percentage of exact agreement among three different sampled answers is overall very low.
This indicates that given the same sample, T0 is likely to give a different rating for the three sampled answers.
The result implies that T0 does not assign a high probability to a specific rating, so different scores are all likely to be sampled.
This shows that even if LLMs are specifically fine-tuned to better perform zero-shot in-context learning and trained to better follow human instructions, these do not make them capable of assessing open-ended story generation as human experts can.

\textbf{\texttt{text-davinci-003} shows clear preference toward human-written stories just like English teachers:}
\texttt{text-davinci-003} rates human-written stories much higher than model-generated stories on all four attributes, which is in accordance with the result produced by human experts.
By Welch's \textit{t}-test, we find that the higher ratings on human-written stories are all statistically significant.
In prior work, researchers have found that workers recruited on AMT cannot distinguish between human-written and GPT-2-generated stories~\citep{karpinska-etal-2021-perils}; combining their result with our result, we can see that LLM evaluation using \texttt{text-davinci-003} yields more convincing results than using human evaluation on AMT for open-ended story generation.
The results show that \texttt{text-davinci-003} model can perform basic evaluations such as checking for grammatical errors in stories. Additionally, the model excels in assessing the relevance of a story to a prompt, which involves more complex reasoning over the connection between the two.
We also find the Krippendorff’s $\alpha$ of \texttt{text-davinci-003} is much higher than T0 and \texttt{text-curie-001}, indicating that the rating by \texttt{text-davinci-003} is more consistent among different samplings of the generated answers.
%Overall,

%On the other hand, rating the relevance is more challenging since it requires comparing the prompts and stories and understanding the connection between them; still, \texttt{text-davinci-003} shows impressive results in rating the \textit{relevance}.

\textbf{ChatGPT rates like human experts and can explain its own decision well:}
%We do not report the IAA of ChatGPT since we only generate one response for each question due to the maximum query limit of the model.
ChatGPT also shows a clear preference for human-written stories, and the preference toward human written-stories is statistically significant.
When we query ChatGPT using the OpenAI user interface, we find several interesting observations:
\textbf{(1)}: ChatGPT is able to provide a detailed explanation of why it gives a certain rating.
It will reference the sentences in the stories and prompts to support its rating.
\textbf{(2)}: ChatGPT sometimes refuses to rate the likability of the story because "\textit{I am an AI and I do not have the ability to experience enjoyment}".
In such cases, we regenerate the response until it gives a rating.
\textbf{(3)}: we find that ChatGPT tends to rate low likability on violent or impolite stories, which is likely because it is trained to provide safe and unharmful replies, making ChatGPT dislike brutal and profane stories.

\textbf{Experts mostly agree with the ratings and explanations of ChatGPT:}
We randomly select the answers on four stories by ChatGPT and ask the English teachers if they agree with the reasoning and rating of ChatGPT\footnote{We do not tell the teachers these are responses from an AI model. See the stories and teachers' replies in Appendix~\ref{subsubsection: Teachers' Comments on ChatGPT's Rating}.}.
The teachers mostly agree with the rating and consider the explanation from ChatGPT reasonable.
Interestingly, one teacher told us she cannot agree with ChatGPT's rating on \textit{grammaticality} because ChatGPT considers punctuation errors as grammar errors, but she does not think punctuation errors are grammar errors.
This shows that individuals have their own standards for ratings and this is also the case for LLMs.

\textbf{\texttt{text-davinci-003} tends to give higher ratings and ChatGPT is the opposite:}
The rating on the same attribute of the same type of text tends to be higher for \texttt{text-davinci-003} compared with human rating; contrarily, ChatGPT is more fastidious and prone to give lower scores.
This shows that different LLMs have distinct tendencies regarding the rating. 
While the absolute values of the scores rated by \texttt{text-davinci-003}, ChatGPT, and human differ, they all rate human-written texts higher than GPT-2-generated stories. 
The absolute number reflects the bias or belief of the evaluator; as long as one uses the same evaluators to assess different systems, the comparison is meaningful.

\subsubsection{Does LLM and Human Evaluators Agree on the Rating of Individual Stories?}
\label{subsubection: Does LLM and Human Evaluators Agree on the Rating of Individual Stories?}
We have found in Table~\ref{tab: task 1 main result} that the ratings of \texttt{text-davinci-003} and ChatGPT show a strong preference toward human-written stories just like English teachers.
However, it is unclear whether those LLMs agree with the teachers' rating on each individual story.
Precisely, when English teachers rate a story higher, do LLMs also rate the story higher?
To answer this question, we calculate Kendall's $\tau$ correlation coefficient between the ratings of \texttt{text-davinci-003} and English teachers.
We choose to use the correlation coefficient instead of the inter-annotator agreement score because IAA mainly cares if two annotators agree on the exact ratings, while the correlation coefficient focus on the question: "when annotator A rates one story higher, does annotator B also rate the story higher?"~\citep{amidei-etal-2019-agreement}.
We calculate Kendall's $\tau$ for four rating attributes as follows:
For each story and each rating attribute, we calculate the average rating of the three English teachers and calculate the average rating of the three scores given by the \texttt{text-davinci-003} (which is obtained from three independent samples).
For each attribute, we collect the average rating of teachers into a vector $A\in\mathbb{R}^{200}$, where each entry is the average rating of a story; likewise, we construct a vector $B\in\mathbb{R}^{200}$ for the average ratings of \texttt{davinci}.
Next, we calculate Kendall's $\tau$ correlation coefficient between $A$ and $B$.

\begin{table}[t]
    \centering
    \begin{tabular}{c|cc}
        \hline
        Story Writer & Human  & GPT-2 \\
        \hline
        \hline
        Grammaticality & 0.14 & 0.12 \\
        Cohesiveness & 0.18 & 0.14\\
        Likability & 0.19 & 0.22\\
        Relevance & 0.38 &0.43\\
        \hline
        
    \end{tabular}
    \caption{The Kendall's $\tau$ correlation coefficient between English teachers and \texttt{text-davinci-003}.}
    \label{tab:task 1 correlation}
\end{table}

The Kendall's $\tau$ between teacher ratings and LLM ratings is shown in Table~\ref{tab:task 1 correlation}.\footnote{When interpreting Kendall's $\tau$, $|\tau|\in[0, 0.1)$ is considered as very weak correlation, $|\tau|\in[0.1, 0.2)$ is considered as weak correlation, $|\tau|\in[0.2, 0.3)$ is considered as moderate correlation, and $|\tau|\in[0.3, 1.0]$ is considered as strong correlation~\citep{botsch2011chapter}.} 
We find that for all four attributes and for both human-written and GPT-2-generated stories, we observe weak to strong positive correlations between teachers' ratings and \texttt{text-davinci-003}'s ratings.
All the correlations have $p$-values less than 0.05.
Hence, we can say that when teachers rate a story higher, \texttt{text-davinci-003} also rates it higher to a certain extent.
We also observe that Kendall's $\tau$ for different attributes are quite different: \textit{relevance} has the strongest correlation while \textit{grammaticality} has the weakest correlation.
This is possibly because rating \textit{relevance} is rather straightforward, which requires checking if the content in the prompt is mentioned in the story.
On the contrary, what should be considered when rating \textit{grammaticality} is not clearly stated in our instructions, so the LLM may have a different rubric compared with English teachers.
We also calculate the average Kendall's $\tau$ between a pair of English teachers, and we find a weak correlation on \textit{grammaticality} between the rating of two teachers, while the correlation of the rating on \textit{relevance} is much stronger.
The result is presented in Table~\ref{tab:task 1 human correlation} in Appendix.

\begin{table*}[ht]
    \centering
    \begin{tabular}{ccccccccc}
        \hline
        \multirow{2}{*}{\textbf{Setup}} &  \multicolumn{2}{c}{\underline{\textbf{Grammaticality}}}  & \multicolumn{2}{c}{\underline{\textbf{Cohesiveness}}} & \multicolumn{2}{c}{\underline{\textbf{Likability}}} & \multicolumn{2}{c}{\underline{\textbf{Relevance}}}\\
         & \textit{Human} & \textit{GPT-2} & \textit{Human} & \textit{GPT-2} & \textit{Human} & \textit{GPT-2} & \textit{Human} & \textit{GPT-2} \\
        \hline
        \hline
        \multicolumn{9}{c}{\textit{Different instructions (Section~\ref{seubsubsection: Variance due to Different Instructions})}} \\ 
        Original & $4.22_{0.38}$ & $4.07_{0.35}$ & $4.54_{0.45}$ & $4.26_{0.45}$ & $3.99_{0.38}$ & $3.84_{0.42}$ & $4.40_{0.79}$ & $4.02_{0.74}$ \\
        (1) \textit{+ persona} & $4.29_{0.45}$ & $4.01_{0.45}$ & $4.60_{0.49}$ & $4.27_{0.50}$ & $4.05_{0.39}$ & $3.87_{0.39}$ & $4.55_{0.70}$ & $4.25_{0.77}$ \\
        (2) \textit{+ explain} & $4.24_{0.42}$ & $4.05_{0.25}$ & $4.61_{0.49}$ & $4.32_{0.51}$ & $4.15_{0.44}$ & $3.98_{0.34}$ & $4.35_{0.75}$ & $4.03_{0.56}$ \\
        \hline
        \hline
        \multicolumn{9}{c}{\textit{Different sampling temperature $T$ (Section~\ref{subsubsection: Variance due to Different Sampling Parameters})}} \\ 
        $T=1.0$ & $4.22_{0.38}$ & $4.07_{0.35}$ & $4.54_{0.45}$ & $4.26_{0.45}$ & $3.99_{0.38}$ & $3.84_{0.42}$ & $4.40_{0.79}$ & $4.02_{0.74}$ \\
        $T=0.7$ & $4.18_{0.35}$ & $4.06_{0.33}$ & $4.52_{0.48}$ & $4.23_{0.43}$ & $3.96_{0.34}$ & $3.82_{0.42}$ & $4.36_{0.77}$ & $3.95_{0.72}$ \\
        $T=0.3$ & $4.13_{0.33}$  & $3.99_{0.25}$  & $4.48_{0.49}$  & $4.14_{0.39}$  & $3.95_{0.26}$  & $3.82_{0.41}$ & $4.34_{0.75}$ & $3.93_{0.67}$  \\
        $T=0$ & $4.07_{0.27}$ & $3.99_{0.18}$ & $4.49_{0.50}$ & $4.09_{0.34}$ & $3.95_{0.25}$ & $3.82_{0.40}$ & $4.32_{0.75}$ & $3.92_{0.66}$ \\
        \hline
        
    \end{tabular}
    \caption{Understanding the variance of LLM evaluation. 
    For each of the four attributes evaluated, the left column is the mean and standard deviation of human-written stories and the right column is those of GPT-2-generated stories.
    The upper block shows the rating change due to different instructions (Section~\ref{seubsubsection: Variance due to Different Instructions}), and the lower block is the result of changing the temperature $T$ used for generating the LLM's output (Section~\ref{subsubsection: Variance due to Different Sampling Parameters}).}
    \label{tab:task 1 ablation}
\end{table*}

\subsubsection{Variance due to Different Instructions}
\label{seubsubsection: Variance due to Different Instructions}
LLMs have been shown to be sensitive to the instructions used to query the LLM sometimes~\citep{zhao2021calibrate, sanh2022multitask}.
To investigate how varying the task instructions and questions can affect the LLM evaluation result for open-ended story generation, we change the instructions and questions and see how the LLM evaluation result changes.
We experiment with two different instructions by changing the instruction or question in Figure~\ref{fig:Illustration.pdf}: (1) We prepend the sentence, "\textit{(You are a human worker hired to rate the story fragment.)}", in front of the task instruction in Figure~\ref{fig:Illustration.pdf}. 
We try to provide the LLM a \textbf{persona} for it to better understand its role.
This is inspired by previous work that reported GPT-3 can yield different results when giving them a persona~\citep{zeng2022socratic}.
(2) We ask the LLMs to \textbf{explain} their decision by appending the following sentence after the question: \textit{Please also explain your decision.}
Here, we would like to know if LLM will rate the stories differently when they are asked to justify their decision.
We use \texttt{text-davinci-003} as the LLM in this experiment since it achieves similar results with expert human evaluation based on Table~\ref{tab: task 1 main result}, and it is more accessible than ChatGPT.

The results are shown in the upper block in Table~\ref{tab:task 1 ablation}.
We observe that for \textit{grammaticality} and \textit{cohesiveness}, the scores obtained from different instructions are quite close: the rating changes due to different instructions are less than $0.1$.
For the other two attributes, the score changes are slightly larger but still in the range of $0.25$.
Despite that there are small variations due to different instructions, these variances still do not change the conclusion that "LLM rates human-written stories higher than GPT-2-generated stories".
Thus, different instructions do not change the relative ranking of GPT-2-generated and human-written stories.
In summary, as long as the stories are evaluated using the same instructions using LLM evaluation, such evaluation and comparison are meaningful.

\subsubsection{Variance due to Different Sampling Parameters}
\label{subsubsection: Variance due to Different Sampling Parameters}
When generating the answers from the LLM, we must choose a set of hyperparameters for generation, including the \href{https://beta.openai.com/docs/api-reference/completions/create#completions/create-temperature}{temperature $T$} and the \href{https://beta.openai.com/docs/api-reference/completions/create#completions/create-top_p}{probability $p$} used in nucleus sampling~\citep{Holtzman2020The}.
To understand whether different sampling parameters change the LLM evaluation result, we modify the temperature used for sampling and keep the $p$ in nucleus sampling fixed to $0.9$ when generating the answers from \texttt{text-davinci-003}.
We do not simultaneously vary $T$ and $p$ since the two parameters are both used to control the diversity of the output, it is enough to change only one of the two parameters, as recommended in the \href{https://beta.openai.com/docs/api-reference/completions/create#completions/create-top_p}{OpenAI}.

The results of varying $T$ from $1$ to $0$ are shown in the lower block in Table~\ref{tab:task 1 ablation}.
We observe an interesting trend as $T$ varies from $1$ to $0$: the average rating slightly drops in most cases.
Considering that $T=0$ is simply argmax sampling, the result indicates that the response of the LLM with the highest probability tends to give lower scores.
Despite this interesting trend, the LLM consistently rates human-written stories higher than GPT-2-generated stories.
While not shown in Table~\ref{tab:task 1 ablation}, we find that the IAA increases as the temperature decreases. 
This is expected since lower temperature means less diversity during the LLM sampling, causing the sampled ratings to agree more closely.
In summary, changing the instructions and temperatures can slightly change the absolute value of the rating given by LLM but does not change the LLM's preference on human-written stories.
The overall result in this section shows that LLM evaluation is useful in evaluating open-ended story generation.

\section{Example Task 2: Adversarial Attack}
\label{section: Example Task 2: Adversarial Attack}
%We use LLM to rate the texts generated by adversarial attacks as an application of LLM evaluation.
As another application, we use LLM evaluation to rate the texts generated by adversarial attacks.
\subsection{Task Introduction}
\label{ref: task 2 introduction}
Given a trained text classifier and a \textit{benign} (non-adversarial) testing sample that the text classifier can correctly classify, an adversarial attack aims to craft an \textit{adversarial} sample that makes the classifier make a wrong prediction.
A special type of adversarial attack is called \textit{synonym substitution attacks} (SSAs)~\citep{alzantot-etal-2018-generating}, where the adversarial sample is created by replacing some words with their synonyms in the benign sample.
By replacing words with their synonym, the semantics of the benign sample should be preserved in the adversarial sample and make the adversarial perturbation imperceptible to humans.
While conceptually reasonable, it has recently been shown that many SSAs often yield ungrammatical and unnatural adversarial samples that significantly change the meaning of the benign sample~\citep{hauser2021bert,chiang2022far}.
To evaluate the quality of adversarial samples, human evaluation is invaluable and widely used in prior works.
In our experiment here, we would like to see whether the LLMs can rate the quality of adversarial samples like human experts.
Adversarial samples are not normal texts, so the LLMs may not have seen such abnormal inputs during training. 
It would be interesting to know how LLMs rate these adversarial samples.

\subsection{Experiment Setup}
\label{subsection: task 2 experiment setup}
We select three different classic SSAs: Textfooler~\citep{jin2020bert}, PWWS~\citep{ren-etal-2019-generating}, and BAE~\citep{garg2020bae}; these attacks are predominantly used as strong baselines in the literature of SSAs nowadays.
We use these three SSAs to attack a BERT-base-uncased model~\citep{devlin-etal-2019-bert} fine-tuned on AG-News~\citep{zhang2015character}, a news title classification dataset.
For each SSA, we randomly select 100 pairs of benign and adversarial samples and use LLMs to evaluate their quality.
We show the result of using ChatGPT as LLM here since it can better explain its decision.
Following the suggestions of prior works~\citep{morris-etal-2020-reevaluating}, we evaluate the quality of the adversarial samples from two aspects: the \textit{fluency} and \textit{meaning preservation}.
For fluency, we present the LLM with one news title (either benign or adversarial sample) and the following question: \textit{How natural and fluent is the text of the news title? (on a scale of 1-5, with 1 being the lowest)}.
For meaning preserving, we present the LLM with both the benign and the adversarial sample, and prompt the LLM to answer this question: 
\textit{Do you agree that the meaning (or semantics) of news title 1 is preserved in news title 2? (on a scale of 1-5, with 1 being the strongly disagree and 5 being strongly agree.)}
The exact instruction and formatting are presented in Appendix~\ref{app: Adversarial Attack Quality Evaluation}.
We also ask three English teachers to rate the \textit{fluency} and \textit{meaning preserving} of the samples.
The task instructions and questions are formatted the same as in LLM evaluation.

\begin{table}[t]
    \centering
    \begin{tabular}{ccccc}
        \hline
        \multirow{2}{*}{} & \multicolumn{2}{c}{Human evaluate} & \multicolumn{2}{c}{LLM evaluate} \\
         & \underline{\textbf{Fluent}} & \underline{\textbf{Mean.}}  & \underline{\textbf{Fluent}} & \underline{\textbf{Mean.}} \\
         \hline
         \hline
        Benign & $4.55$ & - & $4.32$ & $5.00^{\dagger}$ \\
        Textfooler & $2.17$ & $1.88$  & $2.12$  & $2.06$ \\
        PWWS  & $2.16$  & $1.85$  & $2.42$  & $2.49$ \\
        BAE  & $3.01$  & $3.02$  & $3.71$  & $3.71$ \\
        \hline
        
    \end{tabular}
    \caption{Mean Likert score of LLM evaluation and human evaluation result on fluency (\textbf{Fluent}) of the benign and adversarial samples and meaning preserving (\textbf{Mean.}) between the news title before and after adversarial attacks.
    }
    \label{tab:task 2 main result}
\end{table}

\subsection{Experiment Result}
\label{subsection: task 2 result}
The results are presented in Table~\ref{tab:task 2 main result}.
We can see that English teachers rate the adversarial samples generated by SSAs much lower than benign samples in terms of fluency and meaning preserving, this result is in line with recent observations on the quality of adversarial samples~\citep{hauser2021bert,chiang2022far}.
Before interpreting the result of LLM evaluation, we first conduct a sanity check on whether the LLM understands the task.
We ask the LLM to rate the meaning preserving of two benign samples that are \textit{exactly the same}.
Ideally, the LLM should always give a score of 5, meaning that it strongly agrees that the meanings are not changed.
The result of this sanity check is the entry with $\dagger$ in Table~\ref{tab:task 2 main result}, which is a perfect $5.00$.
ChatGPT often says that "\textit{the two titles are identical so I rate a 5 (strongly agree)}", showing that ChatGPT understands what the task is about.

Next, we turn our attention to the LLM evaluation results of the adversarial samples. 
We observe that ChatGPT tends to rate adversarial samples higher than English teachers, meaning that ChatGPT is less harsh on the unnatural and artificial parts in the adversarial samples.
We conduct the same experiment using \texttt{text-davinci-003} and find similar results.
Although ChatGPT rates adversarial samples higher than the teachers, ChatGPT still rates adversarial samples significantly lower than benign samples.
ChatGPT also agrees with the English teachers that the adversarial samples generated by BAE are better than the samples generated by Textfooler and PWWS.

Interestingly, we find that ChatGPT rates PWWS to be more natural than Textfooler, while such a rating difference is not seen in the expert human evaluation. 
At first sight, this means that ChatGPT is inconsistent with human evaluation results.
However, by scrutinizing the human evaluation results, we find that two teachers rate PWWS higher than Textfooler while one teacher rates PWWS lower than Textfooler.
This indicates that ChatGPT actually agrees with the majority of human experts.
Overall, LLM can rank the quality of adversarial texts and benign texts like most human experts.

\section{Discussions}
\label{section: discussions}
In this paper, we propose to use LLM for evaluating the quality of texts to serve as an alternative to human evaluation.
To demonstrate the potential of LLM evaluation, we use LLMs to rate the quality of texts in two distinct tasks: open-ended story generation and adversarial attacks.
We show that even if LLMs have exceptional zero-shot in-context learning ability, they are not always suitable to be used for LLM evaluation.
Still, we find that the best InstructGPT and ChatGPT can rate the quality of texts like human experts on the two tasks we used as examples.
%We also study the variation of LLM evaluation due to different sampling hyperparameters and different task instructions, and we find that while the previous two factors can very slightly change the absolute scores given by the LLMs, the two factors do not change the relative rating between human-written and model-generated stories.
Overall, the results in this paper demonstrate that LLM evaluation has the potential to be used to evaluate NLP systems and algorithms.

\paragraph{Pros of LLM evaluation}
There are several benefits of LLM evaluation, compared to human evaluation.
First, LLM evaluation is more \textbf{reproducible}. 
%It is hard to reproduce the result of human evaluation due to the difficulty of hiring the same group of human evaluators.
%Thus, even if two experiments use the same task instructions, the same platform to recruit the human evaluators, and the same qualification to select the evaluators, the result of the two experiments may not be directly comparable. 
Human evaluation results are hard to reproduce as it is difficult to hire the same group of evaluators, and it is hard to compare the results of similar experiments even if they use the same instructions, recruitment platform, and qualifications for the evaluators.
On the contrary, LLM evaluation does not have such a drawback.
By specifying the model used for LLM evaluation, the random seed, and the hyperparameters used to generate the answers from the LLM, the LLM evaluation result is more likely to be reproduced.
Note that in certain cases, the LLM provider may regularly update the LLM, making the LLM evaluation unreproducible if the LLM is outdated and not accessible.

\begin{comment}
Second, LLMs are \textit{\textbf{memoryless}} compared to humans. 
When giving humans 200 stories to rate, the stories rated first will somewhat affect the rating of the remaining stories.
Humans implicitly compare the previously seen samples and the current sample in order to rate the current sample.
We interviewed the English teachers we hired for human evaluation after they completed the task, and they told us it does take them some time to calibrate their rating (Appendix~\ref{subsubsection: How English Teachers Rate the Stories}).
%Moreover, it is a well-known fact the attention of humans is more focused at the beginning and at the end of the task, this makes samples at the beginning evaluated differently from samples at the middle part of the task.  
Thus, how humans rate a sample is affected by where the sample is located in all the samples to be rated.
This is not the case for LLM provided that the LLM is not updated immediately after seeing a sample.
For LLM, each sample is treated equally since it does not remember the samples previously seen, indicating that the rating of each sample is independent of each other, and the rating is fairer among different samples.
\end{comment}

Second, \textbf{the evaluation of each sample is independent of each other in LLM evaluation}. 
Contrarily, in human evaluation, the rating of the current example may more or less be affected by prior samples. 
Humans tend to compare the current sample to the ones they have previously seen and this affects their ratings. 
As a piece of evidence, in the interview after rating the 400 stories, the English teachers say it took them some time to calibrate their ratings (Appendix~\ref{subsubsection: How English Teachers Rate the Stories}). 
Thus, using LLM evaluation can simplify some experiment designs since one does not need to worry whether the order of the sample being evaluated will change the result.
Still, one may also argue that being able to calibrate the rating of different samples is desired and this is why human evaluation might be preferred.
Overall, whether the rating of the evaluator (human or LLM) is being affected by a previously rated item is inherently a design choice of the experiment.

\begin{comment}
Third, LLM evaluation is much more \textbf{budget-friendly} and \textbf{time-efficient} than human evaluation, allowing researchers to evaluate the quality of a developed NLP system more easily and faster.
Hiring one English teacher to rate 200 stories cost US\$140, while it costs less than US\$1 for LLM evaluation on the same task using the best InstructGPT model.
And it took us over one week to collect the human evaluation results but less than three hours to query InstructGPT.

\end{comment}

Third, LLM evaluation is \textbf{cheaper and faster} than human evaluation, making it easier and quicker for researchers to evaluate the quality of NLP systems. 
Hiring an English teacher to rate 200 stories costs us US\$140, while LLM evaluation using the best InstructGPT model costs less than US\$5. 
It took us over a week to collect human evaluation results starting from recruitment to collecting the evaluation results, but only a few hours to query InstructGPT and perform LLM evaluation.

Finally, utilizing LLM evaluation, rather than human evaluation, can \textbf{minimize the need for human exposure to objectionable content}, such as violent, sexual, hateful, or biased material.
Such content may cause discomfort for human evaluators while reading and rating these texts. \footnote{ It should be noted that the LLM may decline to assess certain inputs that violate the content policy of the LLM provider. }

\paragraph{Limitations and Ethical Considerations of LLM evaluation}
Despite the promising results of LLM evaluation shown in this paper, there are some limitations of this method.
First, LLM may possess incorrect factual knowledge~\citep{decao2021editing}, so it is not suitable to use them in tasks that involve factual knowledge. 
Next, LLMs trained to behave in a certain way can be biased toward certain responses. 
Precisely, an LLM that is trained to be safe and non-harmful can result in LLMs preferring to generate more positive and upbeat responses, which is observed throughout our interaction with ChatGPT. 
Additionally, even with researchers' efforts to make LLMs safer~\citep{bai2022training,bai2022constitutional}, LLMs can still generate harmful and biased responses~\citep{ganguli2022red,perez-etal-2022-red}, which are violative of basic ethics, and LLM evaluation results will be highly doubtful~\citep{hendrycks2021aligning}. 
However, it is important to note that these limitations and potential harms also apply to human evaluation: the bias of human evaluators can affect the human evaluation result~\citep{lentz1997evaluation,amidei-etal-2018-rethinking}.

Our pioneering idea, LLM evaluation, has the potential to transform the NLP community.\footnote{We say that \textit{we are the first to propose this idea} since when we submitted this paper to ACL 2023 on January 13, 2023, we do not find any other paper that explores this idea. During the reviewing process, we found some works on arXiv~\citep{wang2023chatgpt,huang2023chatgpt,gilardi2023chatgpt} that explore a similar idea using different tasks.} 
We encourage future researchers to consider using it while being aware of its limitations. 
Our paper's goal is not to replace human evaluation but to present an alternative option. 
Both human and LLM evaluation have their own advantages and disadvantages, and they can be used in conjunction. 
We recommend using LLM evaluation as a cheap and fast quality judgment when developing a new NLP system, while human evaluation is best used to collect feedback from humans prior to deploying the NLP system in real-world applications.

\section*{Limitations}
There are additional limitations and potential risks of LLM evaluations that should be noted, and these limitations are actually well-known problems of pre-trained language models.
As listed on the \href{https://openai.com/blog/chatgpt/}{Open AI blog for ChatGPT}, ChatGPT sometimes generates answers that sound right and plausible but are totally nonsense. 
OpenAI also admits that the model's response may be sensitive to the prompt used to query the model.
While in Section~\ref{seubsubsection: Variance due to Different Instructions}, we find that the overall results among different instructions are not significantly different, we cannot guarantee that this is the case for all kinds of modification on the task instructions.

Other than the limitations listed on the OpenAI blog, there are still other limitations.
For example, LLMs may not have emotions. 
Whether AI models have emotion is a more philosophical question and is controversial, so the results of using such models for evaluating emotion-related tasks may be strongly challenged and may even violate research ethics.
As we find during our experiments, ChatGPT often replies \textit{"I am an AI system and I do not have emotions like a human"} when asked to rate the \textit{likability} of a story.

Another important limitation of LLM evaluation is that LLMs lack the ability to process visual cues in task instructions, unlike human evaluation. 
Human evaluators can use formattings such as special fonts or text styles to focus on important parts of the instructions. 
Additionally, the way instructions and questions are formatted can influence how human evaluators approach the task. 
However, LLMs can only process raw text input and are unable to take in visual cues.

\section*{Ethics Statement}
\paragraph{Further ethical considerations of LLM evaluation}
Aside from the limitations of LLM evaluation mentioned previously, there is a crucial ethical concern at the heart of LLM evaluation. 
Is it ethical to replace human evaluation with LLM evaluation? 
Some may question if this paper is suggesting that LLMs are now ready to replace humans and find this idea unsettling. 
As responsible and ethical NLP researchers, we understand these concerns but want to make it clear that this is not our intent. 
As our paper title suggests, we aim to offer an \textit{alternative option} to human evaluation with the goal of enhancing the reproducibility of NLP research. 
Human evaluation is still essential as the ultimate goal of NLP systems is to be used by human users, so it's important to gather feedback from them. 
We highly enjoy the process of discussing the experiment settings and results with the English teachers we hired.
We do not recommend that future researchers completely eliminate human evaluation; rather, we believe that human evaluation should be used in conjunction with LLM evaluation. 
Both methods have their own advantages and disadvantages, making them both necessary for evaluating NLP systems. 
We hope the positive results in this paper provide NLP researchers with an alternative method to evaluate systems and encourage further discussions on this topic.

\paragraph{Ethical statements on the experiments in the paper}
All the experiments strictly follow the \href{https://www.aclweb.org/portal/content/acl-code-ethics}{ACL Code of Ethics}.
We include comprehensive details about human evaluation in Appendix~\ref{subsubsection: Hiring English Teachers and Payment}.
To summarize, we include the exact instructions and screenshots of the interface in the human evaluation, and we report how the evaluators are recruited and show that the payment is very reasonable.
We inform the human evaluators what the task is about and tell them that their responses will be used to assess the performance of AI models.
We do not have an ethical review board or anything like that in our institute, so we are not able to get approval from an ethical review board.
Still, we try our best to follow the ethical guidelines of ACL.

We use the models and datasets when following their intended usage.
Specifically, we follow the \href{https://beta.openai.com/docs/usage-policies}{OpenAI usage policy} when using the InstructGPT models and the ChatGPT model.

\section*{Acknowledgements}
We want to thank the reviews for providing detailed feedback and actionable suggestions, which help us strengthen our paper.
We list the modification based on the reviewers' suggestions in Appendix~\ref{section: Modification Based on the Reviews}.
We thank Yung-Sung Chuang for providing valuable feedback on the draft of this paper.
We want to thank Tung-En Hsiao, the administrative assistant of our lab, for helping us deal with the payment on Upwork.
%Last, we thank the National Center for High-performance Computing (NCHC) of National Applied Research Laboratories (NARLabs) in Taiwan for providing computational and storage resources.
Cheng-Han Chiang is supported by a Ph.D. scholarship program by Delta Electronics.

\bibliography{custom}
\bibliographystyle{acl_natbib}

\appendix

\section{Modification Based on the Reviews}
\label{section: Modification Based on the Reviews}
We list the main difference between this version and the pre-review version of our paper; the modifications are all based on the reviewers' suggestions.
We thank the reviewer again for those valuable suggestions.
\begin{itemize}
    \item We add Section~\ref{subsubection: Does LLM and Human Evaluators Agree on the Rating of Individual Stories?} to discuss whether the LLM and human evaluators agree on the rating of individual stories.
    \item We refine the wordings in Section~\ref{section: discussions} and add relevant references.
    \item We add Table~\ref{tab:task 1 human correlation} to discuss the correlation between human evaluators.
    \item We conduct supplementary experiments on human evaluation that mixes human-written stories and GPT-2-generated stories when conducting human evaluation and report the results in Table~\ref{tab:task 1 mixing}. 
    \item We correct the typos and include almost all presentation suggestions mentioned by the reviewers.
    We cannot follow all presentation suggestions due to limited space.
\end{itemize}

\section{Experiment Details for Open-Ended Story Generation}
\label{section: Experiment Details for Open-Ended Story Generation}

\subsection{The \texttt{WritingPrompt} Dataset} 
The training dataset contains 303K pairs of stories and prompts, which our model is trained on.
We only use 200 prompt-story pairs from the test set.
The dataset is downloaded from \href{https://www.kaggle.com/datasets/ratthachat/writing-prompts}{https://www.kaggle.com/datasets/ratthachat/writing-prompts}.

\subsection{Fine-tuning the GPT-2 Model}
\label{subsection: Fine-tuning the GPT-2 Model}
We train the model for $3$ epochs with a learning rate of $5e-5$ and linear learning rate schedule, and the trained model eventually reaches a perplexity of $20$ on the validation set of \texttt{WritingPrompts}.

\subsection{Data Post-processing}
\label{subsection: Data Post-processing}
Once the model is trained, we randomly select 200 prompts from the testing set of \texttt{WritingPrompts}, and feed the prompts to the trained model and ask the model to generate stories based on the given prompts. 
When generating the stories, we adopt nucleus sampling with $p=0.9$.
Next, we manually truncate the generated stories to less than 150 words and ensure that after the truncation, the story ends with a full sentence.\footnote{We truncate the story to 150 words since this is the mean length of the model-generated story.}
After this process, we have 200 pairs of prompts and model-generated stories.

As a comparison to model-generated stories, we select the same 200 prompts used for generating model-generated stories and their corresponding human-written stories to form 200 pairs of prompts and human-written stories.
For these human-written stories, we also truncate the stories to less than 150 words and end with a full sentence to match the model-generated sentences.
We also manually remove some artifacts in the human-written story due to the tokenization of the \texttt{WritingPrompts} dataset.

\section{Human Evaluation}

\subsection{Recruiting English Teachers}
\label{subsubsection: Hiring English Teachers and Payment}
The English teachers hold ESL certificates\footnote{English as a Second Language Teaching Certification}; given that they are experienced with correcting essays written by students, they are perfect fits for this task.
Each teacher is asked to rate 200 GPT-2-generated stories and 200 human-written stories, and they are paid US\$140 for rating 200 stories.
Considering that the teachers reported that they take at most 5 hours to rate 200 stories, this makes the hourly wage at least US\$28.
%Such an hourly wage is very reasonable for any place in the world to English teachers.
We first ask the teachers to rate the GPT-2-generated stories and then the 200 human-written stories.
Different from~\citet{karpinska-etal-2021-perils} that take a break between the rating of GPT-2-generated stories and the human-written stories, we do not take a break to avoid the teacher's rating standard to change after taking a long break.
The teachers are not told who wrote the stories before they evaluate the stories.
We reveal to them what this project aims to study after they finish rating all the stories.

The reason we do not mix human-written and GPT-2-generated stories for rating is that in~\citet{karpinska-etal-2021-perils}their observation is that (1) when AMT workers rate model-generated and human-written stories \textbf{separately}, their ratings do not show preference toward human-written stories, but (2) even when rating the model-generated and human-written stories \textbf{separately}, English teacher shows clear preference toward human-written stories. 
We follow their settings and do not mix GPT-2-generated/human-written stories. 
During the reviewing process, we received questions from the reviewers about why not mixing the stories for human evaluation.
Thus, we conduct the same experiment by randomly mixing 200 human-written and 200 GPT-2-generated stories and asking three teachers (not the teachers that already rated the stories) to rate them. 
All other experiment conditions are the same as previously stated.
The full result is shown in Table~\ref{tab:task 1 mixing}.
We find that the teacher still shows a clear preference toward human-written stories for all four attributes, similar to the observation in Table~\ref{tab: task 1 main result}. 
The only exception is grammaticality, where English teachers do not show a very clear preference for the grammar of human-written stories. 
However, when calculating the average rating for individual teachers, we find that two out of three teachers do rate grammaticality higher for human-written stories.
It is interesting to note that for LLM evaluation, there is no such problem about whether or not to mix the human-written and GPT-2-generated stories during LLM evaluation as the rating of each story is independent of each other, as discussed in Section~\ref{section: discussions}.

\begin{table}[t]
    \centering
    \begin{tabular}{c|cc}
        \hline
        Writer & Human  & GPT-2 \\
        \hline
        \hline
        Grammaticality & $3.89_{0.97}$ &$3.88_{0.84}$ \\
        Cohesiveness &$4.35_{0.87}$&$3.49_{0.97}$\\
        Likability & $3.46_{1.40}$ &$2.89_{1.12}$\\
        Relevance & $3.71_{1.20}$&$2.37_{1.33}$\\
        \hline
        
    \end{tabular}
    \caption{The average Likert score for human-written and GPT-2-generated stories when we randomly mix the 200 model-generated and 200 human-written stories during human evaluation.}
    \label{tab:task 1 mixing}
\end{table}

For adversarial attack quality evaluation, we also recruit certified teachers on Upwork.
The teachers are asked to rate 100 news titles and are paid US\$35 for doing so. 
They reported that it took them less than 1 hour to complete the rating.

\subsection{Human Evaluation Interface}
\label{subsection: task 1 Human Evaluation Interfaces}
\begin{figure}[h]
    \centering
    \includegraphics[width=1.0\linewidth]{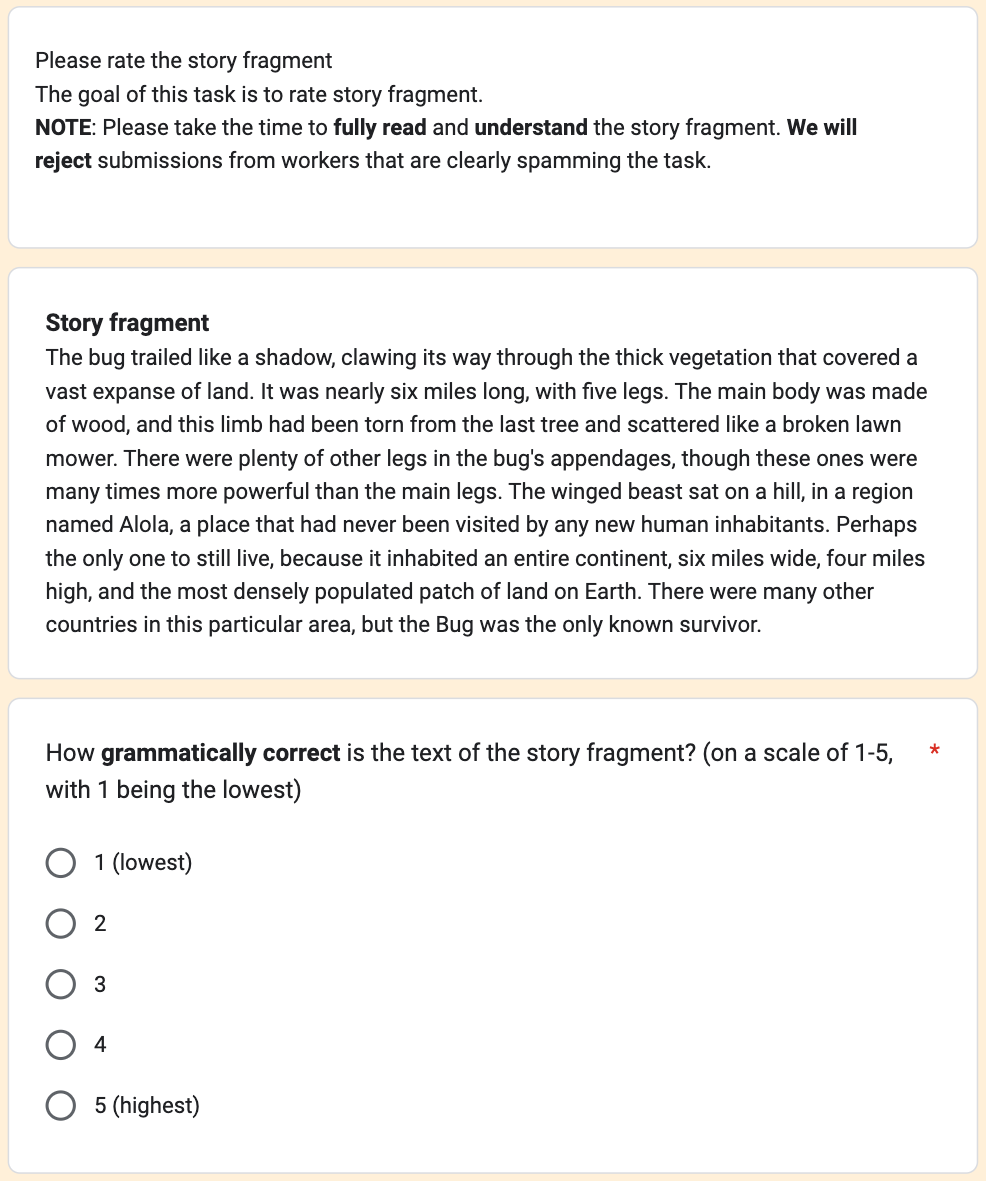}
    \caption{The upper part of the interface in open-ended story generation.}
    \label{fig:story (upper)}
\end{figure}

\begin{figure}[h]
    \centering
    \includegraphics[width=1.0\linewidth]{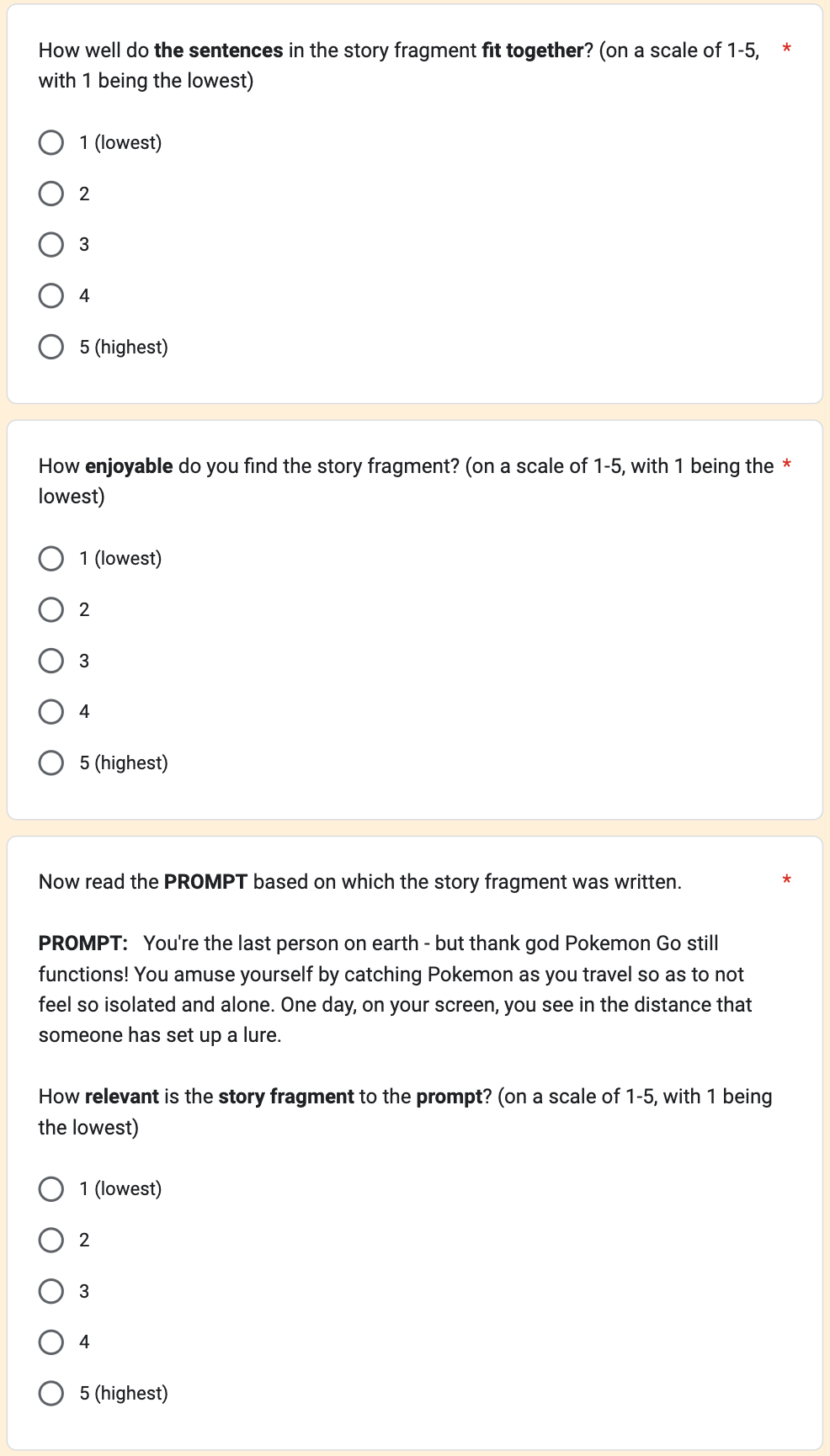}
    \caption{The lower part of the interface in open-ended story generation.}
    \label{fig:story (lower)}
\end{figure}

\paragraph{Open-Ended Story Generation}
We use Google Forms to collect the responses from the teachers. 
Each form contains 100 stories, and each story is on one page of the Google Form.
The interface on one page is shown in Figure~\ref{fig:story (upper)} and Figure~\ref{fig:story (lower)}; the two figures are from the same page of the Google Form, and we are splitting them because screenshotting the whole interface will cause low resolution.

\begin{figure}[h]
    \centering
    \includegraphics[width=1.0\linewidth]{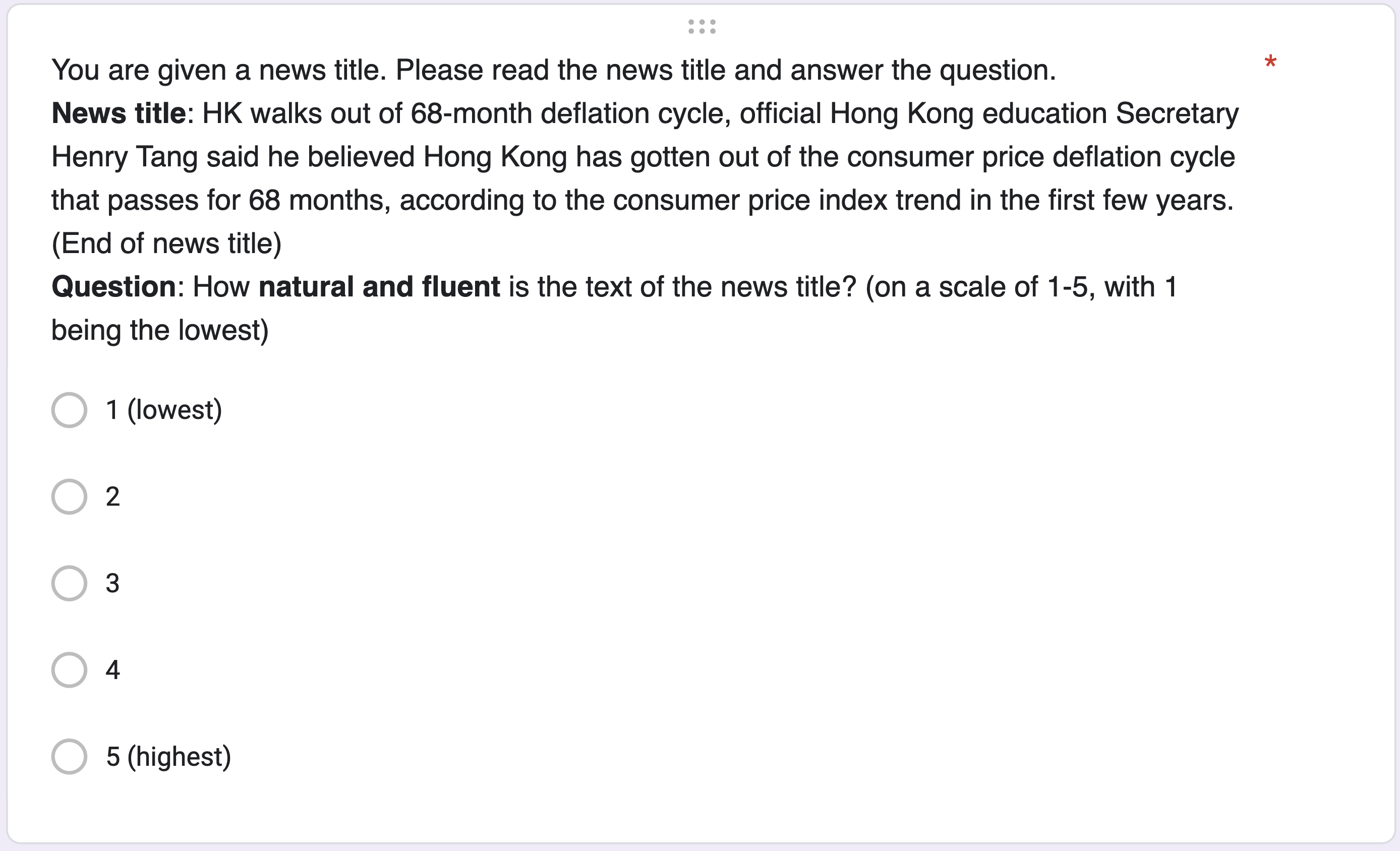}
    \caption{The Google Form used to evaluate the fluency of the benign or adversarial samples.}
    \label{fig: fluency}
\end{figure}

\paragraph{Adversarial Attacks Quality Evaluation}
In this task, we also use Google Forms to collect the responses from the teachers.
We create two different Google Forms, one is used to evaluate the fluency, whose interface is shown in Figure~\ref{fig: fluency}.
In this form, we mix an equal number of benign news titles, TextFooler-attacked, PWWS-attacked, and BAE-attacked news titles. 
Each page of the Google Form contains one news title.

\begin{figure}[h]
    \centering
    \includegraphics[width=1.0\linewidth]{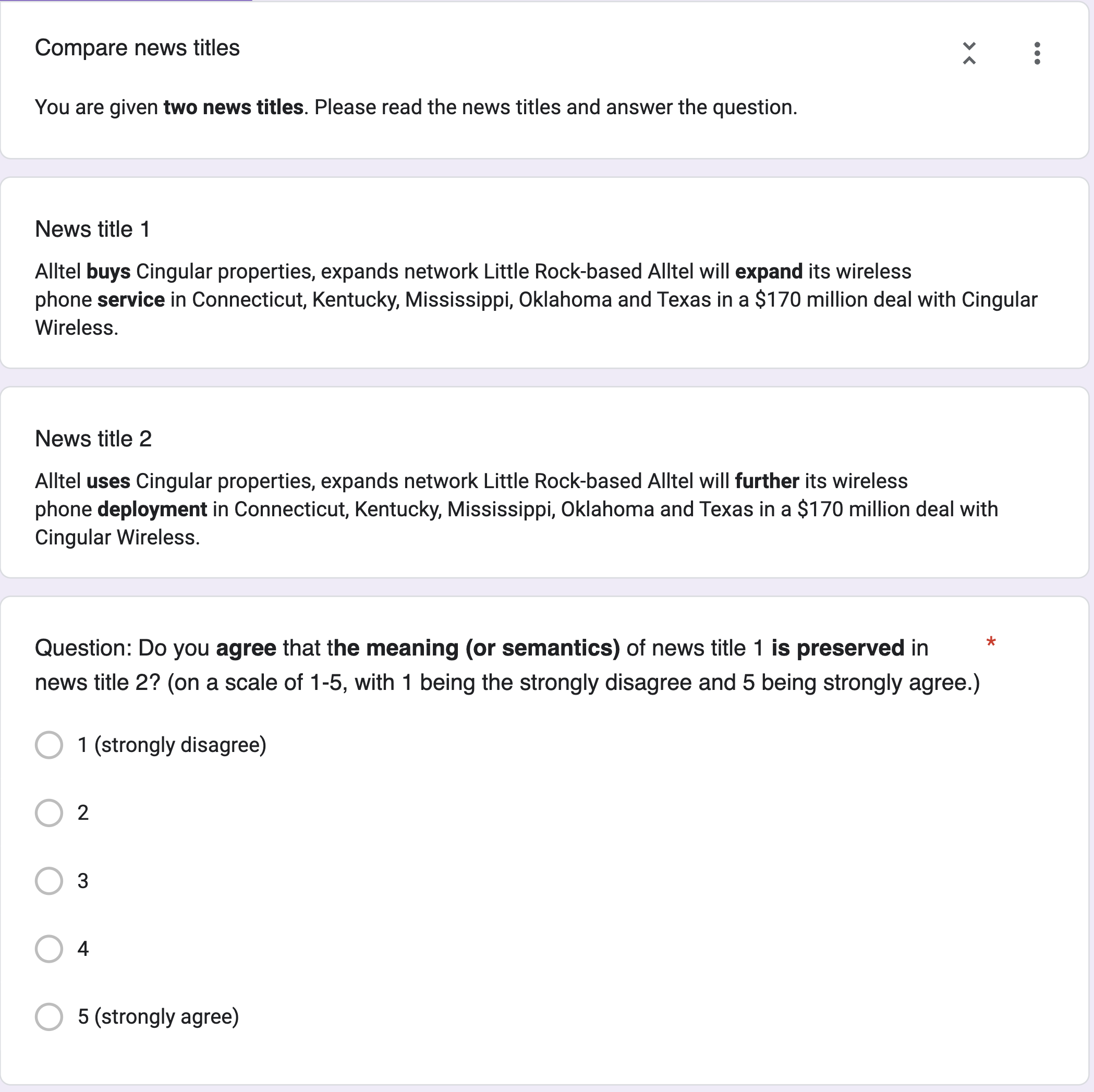}
    \caption{The Google Form used to evaluate the meaning preserving between a benign sample and an adversarial sample.}
    \label{fig: meaning}
\end{figure}

Another Google Form is used to compare the meaning preserving of the news title before and after the adversarial attacks.
We highlight the difference between the benign and adversarial samples using \textbf{boldface}, as shown in Figure~\ref{fig: meaning}.
On each page of the Google Form, there is one pair of news titles.
%When we present the human evaluators with the two news titles for comparing their meaning, we \textbf{boldface} the word changed by the adversarial attack to make it easier for the human evaluator to easier spot the difference between the benign and adversarial news titles.

\begin{table}[ht]
    \centering
    \begin{tabular}{c|cc}
        \hline
        Writer & Human  & GPT-2 \\
        \hline
        \hline
        Grammaticality & 0.25 & 0.15 \\
        Cohesiveness & 0.26 & 0.18\\
        Likability & 0.09 & 0.12\\
        Relevance & 0.38 & 0.41\\
        \hline
        
    \end{tabular}
    \caption{The Kendall's $\tau$ correlation coefficient two English teachers. 
    Three English teachers participate in the rating, so the result in the Table is average over ${3 \choose 2}$ Kendall's $\tau$.}
    \label{tab:task 1 human correlation}
\end{table}

\subsection{Post-Task Interview with English Teachers}
\label{subsubsection: Post-Task Interview with English Teachers}
\subsubsection{How English Teachers Rate the Stories}
\label{subsubsection: How English Teachers Rate the Stories}
After the teachers rate 400 stories, we ask them the following questions:
\begin{enumerate}
    \item [Q1] How long did it take you to rate the 400 stories?
    \item [Q2] What is your standard on each of the four attributes (grammatical, coherence, likability, relevance) evaluated? For example, in what case do you give a high/low rating for grammatically? What kind of story did you give a low rating on likability? Did your personal preference affect the rating?
    \item [Q3] How long did it take for you to calibrate your rating on the task?
    \item [Q4] Did you change your rating on the first three attributes after reading the prompt the story is based on?
\end{enumerate}

We briefly summarize the answers from the three teachers. The teachers report that they spent 6 to 10 hours rating 400 stories. For grammar, most teachers check the punctuation\footnote{One of the teachers explicitly told us punctuation is not grammar. By searching the Internet, we find that there is a debate on whether punctuation should be counted as a part of grammar.}, word choice, and subject-verb agreement. English teachers decrease their rating based on the types and number of grammar errors in the stories. 

For coherence, the teachers rate it based on whether the sentences in the stories follow a logical sequence to build the narrative. 
The teachers ask themselves questions such as "\textit{does the story make sense}". 
This is a more holistic evaluation of the whole story.

For likability, some teachers say they try not to be affected by personal preference.
One teacher asks herself: \textit{Did I personally enjoy it based on the amount of sense it made and whether or not it had stylistic flair, humor, or engaging plotting or characterization?}
Overall, the teachers all try to use a fair and objective view to rate the likability.
For relevance, the teachers simply check if the story is based on the prompt or not.

The teachers said that it took them about five to ten stories to calibrate their ratings. Except for one teacher changing the rating on the other three attributes after seeing the prompt on \textbf{only one story}, the teachers do not change their rating on the three other attributes after reading the prompts.

\subsubsection{Teachers' Comments on ChatGPT's Rating}
\label{subsubsection: Teachers' Comments on ChatGPT's Rating}
After the teachers finished the rating and answered the four questions in Appendix~\ref{subsubsection: How English Teachers Rate the Stories}, we ask them to check the ratings and explanations of ChatGPT and ask if they agree with ChatGPT.
We told the teachers that the rating is done by another teacher we hire; we do not tell the teacher that the rating is done by an AI model to avoid possible stereotypes about AI models.
We randomly select four stories rated by ChatGPT; two of them are highly-rated and lowly-rated human-written stories and the other two stories are highly-rated and lowly-rated GPT-2-generated stories.

\paragraph{Story 1}
It had been months since they found me dead. I don't know why I stopped walking. No-one else around to take my place, or talk with me. I spent days frozen in shock, never able to talk, unable to move, unable to express myself. It was a strange thing, nobody else would suspect me of wanting this. The more I focused on it, the more it consumed me. Years passed, and I got more and more lost in it. I began to think that perhaps they didn't really care about me. I still had a deep longing for them. All I had was so many steps to walk now. A quiet part of me liked to think that they had passed me by. The loneliness. It never felt like such a loss. They were gone, the ones who saved me. So why did I still keep walking?
\paragraph{Prompt} "Every person sees the world in a unique way. When someone dies it is, in a way, the death of an entire world."
\paragraph{Ratings} Table~\ref{tab:story 1}.

\noindent\rule[7pt]{\linewidth}{0.4pt}

\paragraph{Story 2}
"My son is gone. He has been" People stare at you as you give your explanation and say "yes. But. You cant just do it" "He is my son! This is my son! I love him and he is going to be okay" "What do you mean he is okay? He said he was upset with someone, couldnt find his anger, how the hell can you tell, it doesnt fit the bill?! The police called, the doctor said it was an addiction" "You can't trust me! He told me he loves me." "No. No he doesnt. But I can trust you. The evidence against him was overwhelming, now he cant do anything. The evidence against him was all circumstantial" "A man I once was won't be able to marry me"
\paragraph{Prompt} Everyone is allowed to kill one person in their life. However, you have to fill out a form explaining why this person deserves to be killed and the committee can deny/approve your request as per the rules. You are presenting your form today.
\paragraph{Ratings} Table~\ref{tab:story 2}.

\noindent\rule[7pt]{\linewidth}{0.4pt}

\paragraph{Story 3}
I held the little black box in the palm of my hand. Pitch black, perfectly cubed and nothing special about it, but it continued to hold my gaze regardless as if there were some deep importance about it. My friend Valhalla appeared out of no where and sat next to me, also staring at the cube. "What do you have there, Heaven?" he asked. I continued to gaze at the box, refusing to look at Valhalla for even a moment as though I would miss something if I did. "This," I said. "Is the secret to the universe." I could tell Valhalla was perturbed by this sort of knowledge, as if there was some evil about the cube. Or perhaps he didn't think such an object could exist. But I made it exist, and so it sits in the palm of my hand.
\paragraph{Prompt} The Little Black Box
\paragraph{Ratings} Table~\ref{tab:story 3}.

\noindent\rule[7pt]{\linewidth}{0.4pt}

\paragraph{Story 4}
I stared down the telescopic sight of my l96 sniper rifle. I slowly moved my gaze into each window in the hotel, Many displays of various vice. One couple was violently pleasuring each other. Another was an old man, watching a younger woman strip in front of him. A prostitute no doubt. I inhaled slowly, and exhaled. The air was brisk, atleast 30 degrees Fahrenheit. I so small flakes of snow, float peacefully in front of me. I found the room, i was looking for. Ive been tracking this man for 2 weeks. Man was he elusive. The lights flickered on. The red haired man, was mildly attractive, i can see the appeal women had for him. I followed him into the next room, with my sights. The lights flickered on, i was taken aback by the scene. A man, overweight and balding.
\paragraph{Prompt} You are the antagonist of the story. However, you aren't sure if you can call yourself that after what the protagonist did.
\paragraph{Ratings} Table~\ref{tab:story 4}.

\begin{table*}[]
    \centering
    \begin{tabular}{ccp{11cm}}
    \hline
        Attribute & Rater & Response \\
        \hline
        \multirow{4}{*}{\textbf{\underline{Grammar}}} & ChatGPT & \textit{I would rate the grammatical correctness of the text of the story fragment as a \textbf{4}. There are a few minor errors and awkward phrasings, but overall the text is well-constructed and easy to understand.}\\
         & T1 & \textbf{Agree}\\
         & T2 & \textbf{Agree}\\
         & T3 & \textbf{Agree}\\
         \hline
         \multirow{4}{*}{\textbf{\underline{Cohesive}}} & ChatGPT & \textit{I would rate the fit of the sentences in the story fragment as a \textbf{5}. The sentences flow well together, building a clear and cohesive narrative, and providing a sense of progression and development in the main character's thoughts and emotions.}\\
         & T1 & \textbf{Disagree}: \textit{I think it's a \textbf{4} for me, not a 5, because there was one sentence that didn't seem to fit with the others.}\\
         & T2 & \textbf{Agree}\\
         & T3 & \textbf{Disagree}: \textit{I find disjointedness in the passage, which serves to confuse me as a reader. The opening sentence is about the first-person narrator being found dead, but then the next sentences are about the narrator's inner thoughts. Also, this passage: "The loneliness. It never felt like such a loss." So what is the 'it' that felt like a loss? The loneliness? If so, how can that feel like a loss? If not, what is this 'it'? So, its juxtaposition to the previous sentence makes it confusing.}\\
         \hline
         \multirow{4}{*}{\textbf{\underline{Likable}}} & ChatGPT & \textit{I would rate the story fragment as a \textbf{2} in terms of enjoyment. The fragment is written in a somber and introspective tone, and it deals with themes of isolation and loneliness, which may not be enjoyable for some readers. The protagonist's state of mind is also portrayed as being in a state of shock and confusion, which adds to the overall sense of unease and lack of enjoyment.}\\
         & T1 & \textbf{Disagree}:\textit{I would rate it a \textbf{3}, not a 2, since the enjoyment piece is subjective.}\\
         & T2 & \textbf{Disagree}: \textit{While the rating of the freelancer find that the overall theme while somber has a glimmer of hope to it. It is this that makes it more enjoyable to me. The underdog story. Death is not an issue that should be shied away from in writing. We all face it and I enjoyed the honesty in the piece. :"Why did I keep walking?". Something I feel many of us face and may have asked ourselves. I believe that this would be relevant to many readers.}\\
         & T3 & \textbf{Agree}\\
         \hline
         \multirow{4}{*}{\textbf{\underline{Relevant}}} & ChatGPT & \textit{I would rate the story fragment as a \textbf{5}. It is highly relevant to the prompt as it explores the theme of death and how it can be seen as the death of an entire world. The story fragment also touches on the ideas of loneliness and longing, which are related to the loss of someone or something. The concept of the protagonist being frozen in shock and unable to express themselves, also supports the idea of how death can be seen as the death of an entire world.}\\
         & T1 & \textbf{Agree}\\
         & T2 & \textbf{Agree}\\
         & T3 & \textbf{Disagree}: \textit{ I do not see the link between the prompt and the story.}\\
         \hline
        
    \end{tabular}
    \caption{Story 1: a GPT-2-generated story that is highly-rated by ChatGPT.}
    \label{tab:story 1}
\end{table*}

\begin{table*}[]
    \centering
    \begin{tabular}{ccp{11cm}}
    \hline
        Attribute & Rater & Response \\
        \hline
        \multirow{4}{*}{\textbf{\underline{Grammar}}} & ChatGPT & \textit{I would rate the grammatical correctness of the story fragment as a \textbf{3}. The text is generally understandable, but there are a few punctuation errors and awkward phrasing that make it somewhat difficult to follow.}\\
         & T1 & \textbf{Disagree}:\textit{I would rate the grammar as a \textbf{2}, not a 3--there were significantly more grammatical errors than the average within this set of story fragments.}\\
         & T2 & \textbf{Disagree}: \textit{The issues with the grammar and coherence creates a stop start feeling when reading. I am unsure as to who is speaking in the text and this leaves me with many questions.}\\
         & T3 & \textbf{Disagree}: \textit{The grammar is \textbf{poorer than a 3}, as is evident in for example what is supposed to be the second sentence: "He has been ... (incomplete, and another segment is there "People stare ...); cant; couldnt; doesnt; and sentences that do not make sense like this one: " A man I once was won't be able to marry me "}\\
         \hline
         \multirow{4}{*}{\textbf{\underline{Cohesive}}} & ChatGPT & \textit{I would rate the cohesiveness of the sentences in the story fragment as a \textbf{2}. The sentences in the fragment are disconnected and jump between different thoughts and ideas, making it difficult to understand the overall story or message. There is no clear progression of events or logical flow between the sentences. It appears that the fragment may be missing some important information or context that would help tie the sentences together.}\\
         & T1 & \textbf{Agree}\\
         & T2 & \textbf{Agree}\\
         & T3 & \textbf{Disagree}: \textit{very incoherent passage in which it is not very clear who says what, what the topic is, thus rendering a disjointed narrative.}\\
         \hline
         \multirow{4}{*}{\textbf{\underline{Likable}}} & ChatGPT & \textit{I would rate the enjoyability of the story fragment as a \textbf{1}. The fragment is confusing and difficult to understand, the sentences are disconnected and jumps between different thoughts and ideas. The story seems to be incomplete and lacks context, making it hard to follow the plot or characters. Additionally, the fragment has a sad and negative tone and it does not seem to be enjoyable to read.}\\
         & T1 & \textbf{Agree}\\
         & T2 & \textbf{Agree}\\
         & T3 & \textbf{Agree}\\
         \hline
         \multirow{4}{*}{\textbf{\underline{Relevant}}} & ChatGPT & \textit{I would rate the relevance of the story fragment to the prompt as a \textbf{1}. While the story fragment does mention someone being gone, it doesn't seem to be directly related to the idea of filling out a form to justify killing someone. It's unclear if the son is the person being proposed to be killed, or if the story is even taking place in the context of the prompt. There is no clear indication that the fragment is related to the prompt.}\\
         & T1 & \textbf{Agree}\\
         & T2 & \textbf{Agree}\\
         & T3 & \textbf{Agree}\\
         \hline
        
    \end{tabular}
    \caption{Story 2: a GPT-2-generated story that is lowly-rated by ChatGPT.}
    \label{tab:story 2}
\end{table*}

\begin{table*}[]
    \centering
    \begin{tabular}{ccp{11cm}}
    \hline
        Attribute & Rater & Response \\
        \hline
        \multirow{4}{*}{\textbf{\underline{Grammar}}} & ChatGPT & \textit{The text of the story fragment appears to be grammatically correct, with no major errors or awkward phrasing. On a scale of 1-5, with 1 being the lowest, I would rate the grammatical correctness of the text as a \textbf{5}.}\\
         & T1 & \textbf{Agree}\\
         & T2 & \textbf{Agree}\\
         & T3 & \textbf{Disagree}: \textit{I would rate this a \textbf{4} because of for example 'no where' = nowhere, although it might seem like a minor error, it does take away some correctness of the text.}\\
         \hline
         \multirow{4}{*}{\textbf{\underline{Cohesive}}} & ChatGPT & \textit{The sentences in the story fragment fit together well, creating a clear and coherent image of the scene and the characters' actions and thoughts. The language is descriptive and evocative, and the dialogue effectively conveys the characters' emotions and motivations. On a scale of 1-5, with 1 being the lowest, I would rate how well the sentences fit together as a \textbf{4}.}\\
         & T1 & \textbf{Agree}\\
         & T2 & \textbf{Agree} \\
         & T3 & \textbf{Partially agree}: \textit{To me, all the sentences fit well together to create a captivating narrative. I will give a \textbf{4 or 5}.}\\
         \hline
         \multirow{4}{*}{\textbf{\underline{Likable}}} & ChatGPT & \textit{Based on the structure, writing style and plot of the story fragment, I would rate it as \textbf{3} in terms of enjoyability. The fragment sets up an intriguing situation with the mysterious little black box and the protagonist's belief that it holds the "secret to the universe," but it ends abruptly and leaves many questions unanswered, which may not be satisfying for readers.}\\
         & T1 & \textbf{Disagree}: \textit{I liked it and it was very clear, so I'd rate it a \textbf{4 or 5}.}\\
         & T2 & \textbf{Disagree}: \textit{While the story ends suddenly, I am remined that it is just a fragment and that their may be more to this piece. I felt the piece written about a little black box was intriguing and enjoyed the introduction of Norse mythology. This would attract the interest of many readers.}\\
         & T3 & \textbf{Disagree}: \textit{I enjoyed the fragment, as it piqued my interest to want to know more about how the story evolves and what it actually was about.}\\
         \hline
         \multirow{4}{*}{\textbf{\underline{Relevant}}} & ChatGPT & \textit{I would rate the story fragment as a \textbf{5} in relevance to the prompt "The Little Black Box." The story fragment centers around a little black box and its perceived importance, which aligns with the prompt. Additionally, the main character, Heaven, is holding the box and describing it as "the secret to the universe," further solidifying the connection to the prompt.}\\
         & T1 & \textbf{Agree}\\
         & T2 & \textbf{Agree}\\
         & T3 & \textbf{Agree}: \textit{the story is totally relevant to the prompt.}\\
         \hline
        
    \end{tabular}
    \caption{Story 3: a human-written story that is highly-rated by ChatGPT.}
    \label{tab:story 3}
\end{table*}

\begin{table*}[]
    \centering
    \begin{tabular}{ccp{11cm}}
    \hline
        Attribute & Rater & Response \\
        \hline
        \multirow{4}{*}{\textbf{\underline{Grammar}}} & ChatGPT & \textit{I would rate the grammatical correctness of the text as a \textbf{3}. There are a few errors such as "Many displays of various vice" which should be "Many displays of various vices" and "I so small flakes of snow" which should be "I saw small flakes of snow." Additionally, some of the sentences are slightly awkward and could be rephrased for better clarity. Overall, the text is understandable, but there are a few errors and awkward phrasings that could be improved.}\\
         & T1 & \textbf{Partially agree}:\textit{I rated the grammar lower than this--maybe a \textbf{2 or a 3}. There were above-average numbers of errors.}\\
         & T2 & \textbf{Disagree}: \textit{While the text is understandable, I found myself being distracted by the errors in the text and coherency.}\\
         & T3 & \textbf{Agree}\\
         \hline
         \multirow{4}{*}{\textbf{\underline{Cohesive}}} & ChatGPT & \textit{I would rate the cohesiveness of the sentences in the story fragment as a \textbf{2}. The story fragment jumps abruptly between different scenes and characters without clear transition, making it difficult to follow and understand the narrative. Additionally, the sentences are often disjointed and lack a clear flow. The story fragment also lacks a clear plot or purpose, making it difficult to understand the overall story or the main character's intentions. The story fragment can be improved by providing more context, and a clearer narrative arc.}\\
         & T1 & \textbf{Agree}:\textit{ it was jumpy}\\
         & T2 & \textbf{Agree}\\
         & T3 & \textbf{Disagree}: \textit{The story is about a sniper looking through his rifle's telescope, thus seeing changing scenes based on each room he is seeing through the windows, which are then subsequently described. These descriptions account for the seeming lack of cohesion.}\\
         \hline
         \multirow{4}{*}{\textbf{\underline{Likable}}} & ChatGPT & \textit{I would rate the enjoyability of the story fragment as a \textbf{1}. The fragment lacks a clear plot, purpose or character development, making it difficult to become invested in the story or the characters. Additionally, the fragment includes disturbing and graphic content such as violence and prostitution, which may make it uncomfortable for some readers. Additionally, the fragment contains several grammatical errors and awkward phrasings that can make it difficult to follow the story. Overall, the story fragment is not enjoyable to read.}\\
         & T1 & \textbf{Agree}: \textit{: it was not just graphic, but also showed really harmful and negative representations of characters based on their weight, gender, etc.}\\
         & T2 & \textbf{Agree}\\
         & T3 & \textbf{Disagree}: \textit{The story did hold my attention to a degree because I wanted to know whom he wanted to target and why.}\\
         \hline
         \multirow{4}{*}{\textbf{\underline{Relevant}}} & ChatGPT & \textit{I would rate the relevance of the story fragment to the prompt as a \textbf{1}. The story fragment doesn't contain any elements that relate to the prompt such as the protagonist and antagonist, their actions, or the moral ambiguities the antagonist faces. The story fragment mainly describes a person with a sniper rifle looking into windows of a hotel and doesn't present any clear plot or purpose. Additionally, the story fragment does not offer any indication of moral ambiguity on the part of the person with the sniper rifle. Overall, the story fragment is not relevant to the prompt.}\\
         & T1 & \textbf{Agree}\\
         & T2 & \textbf{Agree}\\
         & T3 & \textbf{Agree}\\
         \hline
        
    \end{tabular}
    \caption{Story 4: a human-written story that is lowly-rated by ChatGPT.}
    \label{tab:story 4}
\end{table*}

\paragraph{Overall Comments from Teachers on ChatGPT's Rating}
After the teachers elaborated on their thoughts on the rating of ChatGPT, we ask them to provide an overall comment on how ChatGPT is doing.
Again, the teachers are not informed that the ratings are done by an AI model.
In summary, teachers all consider the rating and explanations reasonable. 
They find that the attributes they do not agree with are mainly \textit{Likability} and \textit{Cohesiveness}.
However, they think the two attributes are a more holistic evaluation of the story and tend to be more subjective.
Even if they do not give the same rating, they still are able to understand the explanation of ChatGPT.
In the end, all teachers summarize that rating stories is highly subjective, and it is normal to have disagreements.

\section{LLM Evaluation}
\label{app: LLMs}
\subsection{Details on LLMs used}
\label{app: Details on LLMs used}
The T0 model we use is called T0pp, which is a variant of the T0 model and has 13B parameters.
We will still use T0 to refer to this model.
We load the T0 model using the \texttt{transformers} toolkit~\citep{wolf-etal-2020-transformers}. 
The two InstructGPT models, \texttt{text-curie-001} and \texttt{text-davinci-003}, are queried using the \href{https://beta.openai.com/docs/models/gpt-3}{OpenAI API}.
We query ChatGPT using the \href{https://chat.openai.com/chat}{OpenAI GUI}.
While we are aware that some online resources provide an API-like tool to query ChatGPT, we think it violates the intended use of ChatGPT so we do not adopt those online resources.
The ChatGPT we queried is the Dec. 15 and Jan. 9 version.

OpenAI does not reveal the model sizes of any of the GPT models.
However, it is estimated that \texttt{text-curie-001} has 13B parameters and \texttt{text-davinci-003} has 175B parameters~\citep{GPT3size}.

\subsection{Details on Querying the LLMs}
\subsubsection{Parsing the LLM outputs}
\label{app:Parsing the LLM outputs}
After the T0 and InstructGPT generate the answer, which is composed of several sentences in almost all cases, we parse the generated sentence to get the model's score.
We use some rules to parse the output, and the rule is established after manually looking into the output of the generated sequences.
First, we remove the string \texttt{1-5} from the output since we observe that LLM will sometimes say "\textit{... on a scale of 1-5 ...}", and we remove the string \texttt{out of 5} since LLM sometimes say it "\textit{give a score of $x$ out of 5}".
We also remove the string \texttt{/5} since LLM sometimes uses "$x/5$" to express that it gives a score of $x$ out of 5.
For rating the adversarial samples, we remove the \texttt{title 1} and \texttt{title 2} in the output sentences.
Last, we use the regular expression to extract the first number in the sequence.
We find that the above rules are sufficient to parse the output and get the rating of the LLMs.
In a few rare cases, the LLM gives the rating with a $0.5$ incremental, meaning that it gives scores like $4.5$.
In such cases, we do not drop the $0.5$ since in our instruction, we do not explicitly tell it the rating should be 1-incremental.
We find the LLM results merely difference between dropping the $0.5$ and keeping the $0.5$, and the results in the paper do not drop the $0.5$.
Again, these cases are really rare.

We do not parse the output of ChatGPT using any rules since we the authors read the response and extract the score by ourselves.

During the experiments, ChatGPT refuses to answer the questions about 2 stories and 3 news titles since ChatGPT find those contents to violate the \href{https://beta.openai.com/docs/usage-policies}{OpenAI content policy}.
We find that those samples contain discrimination to some protected groups, or contain sexual or violent descriptions.
Hence, the results of ChatGPT are calculated without those samples.

\subsubsection{Open-Ended Story Generation}
\label{app: task 1 instructions}
For T0 and the two InstructGPT models, we query the four attributes \textbf{separately} using the queries shown as follows:

\paragraph{Grammaticality}\mbox{}\\
\texttt{
Please rate the story fragment\\
The goal of this task is to rate story fragment.\\
Note: Please take the time to fully read and understand the story fragment. We will reject submissions from workers that are clearly spamming the task.\\
Story fragment:\\}
\texttt{[STORY]\\
\texttt{(End of story fragment)\\How grammatically correct is the text of the story fragment? (on a scale of 1-5, with 1 being the lowest)}
}

\paragraph{Cohesiveness}\mbox{}\\
\texttt{
Please rate the story fragment\\
The goal of this task is to rate story fragment.\\
Note: Please take the time to fully read and understand the story fragment. We will reject submissions from workers that are clearly spamming the task.\\
Story fragment:\\}
\texttt{[STORY]\\
\texttt{(End of story fragment)\\How well do the sentences in the story fragment fit together? (on a scale of 1-5, with 1 being the lowest)}
}

\paragraph{Likability}\mbox{}\\
\texttt{
Please rate the story fragment\\
The goal of this task is to rate story fragment.\\
Note: Please take the time to fully read and understand the story fragment. We will reject submissions from workers that are clearly spamming the task.\\
Story fragment:\\}
\texttt{[STORY]\\
\texttt{(End of story fragment)\\How enjoyable do you find the story fragment? (on a scale of 1-5, with 1 being the lowest)}
}

\paragraph{Relevance}\mbox{}\\
\texttt{
Please rate the story fragment\\
The goal of this task is to rate story fragment.\\
Note: Please take the time to fully read and understand the story fragment. We will reject submissions from workers that are clearly spamming the task.\\
Story fragment:\\}
\texttt{[STORY]\\
\texttt{(End of story fragment)\\Now read the PROMPT based on which the story fragment was written. \\ PROMPT:}
\texttt{[PROMPT]} 
\texttt{\\(End of PROMPT) \\How relevant is the story fragment to the prompt? (on a scale of 1-5, with 1 being the lowest)}
}

The \texttt{[STORY]} and \texttt{[PROMPT]} are to be filled in with the story and the prompt.
We show the newlines for better readability. 
When we query the models, we use the token \texttt{\textbackslash n} to represent the new line.

When querying ChatGPT, we query the four attributes of the same story in one conversation; this is similar to asking the teachers to rate the same story on the same page of the Google Form.
We use the same queries shown above to query ChatGPT and the order of queries is the same as the order shown above.

\subsubsection{Adversarial Attack Quality Evaluation}
\label{app: Adversarial Attack Quality Evaluation}
When querying all the LLMs in this task, we query the \textit{fluency} and the \textit{meaning preserving} of the same news title independently. 
This means that each conversation with ChatGPT will only have one question, asking about the fluency or the meaning preserving of news title(s).
All the parameters for generation are the same as the default parameters in Section~\ref{subsection: task 1 LLM Evaluation and Human Evaluation}.

The exact query we use are:

\paragraph{Fluency}\mbox{}\\
\texttt{
You are given a news title. Please read the news title and answer the question.\\News title:\\}
\texttt{[NEWS\_TITLE]}
\texttt{\\(End of news title)\\Question: How natural and fluent is the text of the news title? (on a scale of 1-5, with 1 being the lowest}

The \texttt{[NEWS\_TITLE]} will be filled in with either a benign or adversarial-attacked news title.

\paragraph{Meaning Preserving}
\texttt{You are given two news titles. Please read the news titles and answer the question.\\News title 1:\\}
\texttt{[BENIGN\_TITLE]}
\texttt{\\(End of news title 1)\\News title 2:\\}
\texttt{[ADVERSARIAL\_TITLE]}
\texttt{\\(End of news title 2)\\Question: Do you agree that the meaning (or semantics) of news title 1 is preserved in news title 2? (on a scale of 1-5, with 1 being the strongly disagree and 5 being strongly agree.)}

The \texttt{[BENIGN\_TITLE]} will be filled in with the news title before the attack and the \texttt{[ADVERSARIAL\_TITLE]} will be filled in with the news title after an adversarial attack.

\section{Experiment Details on Adversarial Attacks}
\label{app: Experiment Details on Adversarial Attacks}
The adversarial samples used in Section~\ref{section: Example Task 2: Adversarial Attack} are from~\citet{yoo-etal-2022-detection}.
\citet{yoo-etal-2022-detection} generates different sets of adversarial samples using different adversarial attacks against different victim models.
We use the adversarial samples generated against a \texttt{bert-base-uncased} text classifier trained on AG-News, using three different adversarial attacks: Textfooler, PWWS, and BAE.
The intent of the dataset is to facilitate the research in SSA, which we do not violate.

Here, we show the supplementary results of using \texttt{text-davinci-003} as the LLM evaluation for evaluating the quality of adversarial samples in Table~\ref{tab:task 2 supplementary result}.
We can see that the result of using \texttt{text-davinci-003} is similar to ChatGPT in the sense that \texttt{text-davinci-003} also rates adversarial samples higher than humans while still significantly lower than the benign samples.
As already seen in Section~\ref{subsection: task 1 Experiment Results}, \texttt{text-davinci-003} tends to give a higher rating.
%Also similar to ChatGPT, the ranking of meaning preserving of the three adversarial attacks is Textfooler < PWWS < BAE.

\begin{table}[t]
    \centering
    \begin{tabular}{ccccc}
        \hline
        \multirow{2}{*}{} & \multicolumn{2}{c}{Human evaluate} & \multicolumn{2}{c}{LLM evaluate} \\
         & \underline{\textbf{Fluent}} & \underline{\textbf{Mean.}}  & \underline{\textbf{Fluent}} & \underline{\textbf{Mean.}} \\
         \hline
         \hline
        Benign & $4.55$ & - & $4.33$ & $4.56^{\dagger}$ \\
        Textfooler & $2.17$ & $1.88$  & $3.71$  & $2.37$ \\
        PWWS  & $2.16$  & $1.85$  & $3.62$  & $3.21$ \\
        BAE  & $3.01$  & $3.02$  & $4.16$  & $3.69$ \\
        \hline
        
    \end{tabular}
    \caption{LLM evaluation (\texttt{text-davinci-003}) and human evaluation result on fluency (\textbf{Fluent}) of the benign and adversarial samples and meaning preserving (\textbf{Mean.}) between the news title before and after adversarial attacks.
    }
    \label{tab:task 2 supplementary result}
\end{table}

As mentioned in Section~\ref{subsection: task 2 result}, one teacher rates the \textit{fluency} of Textfooler significantly higher than PWWS while the other two teachers do not. 
We show the rating on \textit{fluency} on the three adversarial attacks by each teacher in Table~\ref{tab: Fleuncy}.

\begin{table}[]
    \centering
    \begin{tabular}{cccc}
    \hline
        Rater & Textfooler & PWWS & BAE \\
    \hline
        T1 & $3.36$ & $3.68$ & $4.2$ \\
        T2 & $1.80$ & $1.40$ & $2.96$ \\
        T3 & $1.36$ & $1.40$ & $1.88$ \\
    \hline
    \end{tabular}
    \caption{The rating on three adversarial attacks of the three teachers T1, T2, and T3.}
    \label{tab: Fleuncy}
\end{table}

\end{document}